\documentclass{article} %
\usepackage{iclr2027_conference,times}

\usepackage{amsmath,amsfonts}

\def\eqref#1{equation~\ref{#1}}

\def\1{\bm{1}}

\DeclareMathAlphabet{\mathsfit}{\encodingdefault}{\sfdefault}{m}{sl}
\SetMathAlphabet{\mathsfit}{bold}{\encodingdefault}{\sfdefault}{bx}{n}

\usepackage{graphicx}
\usepackage{hyperref}
\hypersetup{hidelinks}
\usepackage{url}
\usepackage{booktabs}
\usepackage{xspace}

\newcommand{\placeholder}[1]{\textit{[Placeholder: #1]}}
\newcommand{\figureplaceholder}[2]{%
  \fbox{\parbox[c][#1][c]{\dimexpr\linewidth-2\fboxsep-2\fboxrule\relax}{%
    \centering\placeholder{#2}}}}
\newcommand{\method}{RoboForge\xspace}

\title{Agentic Robot Creation through Generation, Verification, and Revision}
\author{%
  Xuebing Li\textsuperscript{1},
  Yibo He\textsuperscript{1},
  Sitong Liu\textsuperscript{1},
  Ran Ji\textsuperscript{1},
  Jiayuan Gu\textsuperscript{1}\\
  {\normalfont\textsuperscript{1}ShanghaiTech University}%
}
\iclrfinalcopy %

\usepackage{array}
\usepackage{tabularx}
\usepackage{booktabs}
\usepackage{longtable}
\usepackage{placeins}

\begin{document}
\maketitle
\fancyhead{} %
\renewcommand{\headrulewidth}{0pt}

\begin{abstract}
Can an AI agent turn high-level user intent into a functioning robot and teach it how to behave? We study this question from two complementary aspects: embodiment and behavior creation. For embodiment creation, we investigate how well current multimodal models capture mechanical requirements, and introduce computational metrics for evaluating generated designs before fabrication. For behavior creation, we present \textbf{\method}, an agentic framework that turns task specifications into executable policies through motion generation, policy learning, and execution-grounded revision. Given a robot embodiment and task, the agent generates a reference motion, configures the learning objective, and trains a policy. Crucially, motion remains in the loop: rollout feedback can revise the reference motion, reward, or both. We evaluate a generated embodiment across multiple MLLMs and behavior creation across four diverse robots against reward-refinement and one-shot motion baselines. Our computational metrics expose recurring mechanical failures in generated designs, while behavior experiments show that execution-grounded refinement improves task-level behavior and execution quality. Finally, we fabricate an agent-generated Gorilla and directly deploy learned behaviors on the physical robot.
Project page: \url{https://roboforge-project.github.io/}
\end{abstract}

\section{Introduction}
\label{sec:introduction}

\smallskip
\noindent{\itshape ``Design is not just what it looks like and feels like; design is how it works.'' --- Steve Jobs}
\smallskip

Small expressive robots such as Microduck and recent character robots~\citep{grandia2025design,muller2026olaf} suggest a future in which people can personalize not only how a robot looks, but also how it moves and behaves. Yet creating such a robot today still demands substantial expertise. A high-level concept must be translated into a mechanical structure, suitable joints and actuators, a valid simulation model, desired motions, and ultimately a control policy capable of executing them. These stages typically require mechanical designers, animators, and control engineers to iterate together. With increasingly capable multimodal foundation models, we ask: \textbf{can robot creation become an agentic process, where users describe what they want and agents carry out much of the engineering required to make it physically realizable?}

Recent work has begun to automate individual stages of this process. Foundation models have been used to generate robot morphologies and articulated designs~\citep{ringel2025text2robot,sontakke2026robotdesigngpt,wang2025kinematify}. In parallel, LLMs can synthesize and iteratively refine rewards for reinforcement learning~\citep{ma2024eureka,lee2026rda}, while MimicAgent generates reference motions for downstream policy learning~\citep{nayak2026mimicagentquadrupedskillsprompttotrajectory}. Together, these advances suggest that much of the robot-creation pipeline may become amenable to agentic automation. Connecting these pieces, however, reveals a fundamental gap: \textbf{a robot that looks plausible is not necessarily buildable, and a motion that looks plausible is not necessarily executable}.

We study this gap from two aspects of robot creation. 
For \textbf{embodiment creation}, rather than introducing another morphology generator, we ask how well current multimodal large language models (MLLMs) understand the mechanical requirements that matter before fabrication.
A visually convincing CAD model may still contain disconnected structures, insufficient actuator clearance, self-collisions, or unsuitable contact geometry. 
We therefore identify a set of mechanically relevant requirements and introduce multi-faceted computational metrics to quantify them before fabrication. Using these metrics, we systematically compare different MLLMs, characterize the mechanical knowledge they already possess, and expose recurring failure modes in their generated designs. Importantly, we have fabricated an agent-generated robot \emph{Gorilla} and demonstrate that designs passing these checks can be assembled and operated as physical systems.

For \textbf{behavior creation}, physical feasibility is harder to assess from generation alone. A motion may appear natural but remain incompatible with a particular embodiment because of its timing, pose, contact sequence or joint excursion. Such incompatibilities often become apparent only when a policy attempts to execute the motion.
This observation motivates \textbf{\method}, an agentic framework for turning desired behaviors into executable robot policies through \textbf{generation, physical execution, and revision}. Given a robot embodiment and a task description, the agent generates a reference motion and a learning objective, trains a policy, and inspects the resulting rollouts. Execution therefore serves as a physical verification signal rather than merely the endpoint of learning. When the policy fails, the agent reasons about whether the failure stems from \emph{what behavior is being requested}---such as motion timing, target pose, contact sequence, or trajectory---or from \emph{how the behavior is being learned}---such as reward terms and their relative weights. It then revises the corresponding component and retrains the policy. This closes a motion--policy feedback loop in which learning outcomes expose deficiencies in generated motions, while generated motions continue to provide structured guidance for learning.

We evaluate behavior creation across four diverse embodiments, including existing platforms such as Unitree Go2 and Microduck as well as agent-generated robots, and compare \method with representative reward-refinement and one-shot motion-generation approaches. \method leads to policies that are better aligned with the intended task, as evidenced by both user preference studies and complementary quantitative metrics. Besides, we analyze the intermediate failures and revisions throughout the process to understand how execution feedback drives improvement. We further deploy the resulting policies on the physical robot we fabricate, demonstrating behaviors that transfer from high-level user intent to real-world execution.

\section{Related Work}

\textbf{Automated Robot Generation.}
Automating robot design has progressed from morphology search and modular optimization toward agent-driven generation. Prior methods translate high-level specifications into robot morphologies through language-guided search or generative design~\citep{ma2024exploration,ringel2025text2robot,song2025laser,qiao2025lang2morph}. Recent methods further generate articulated robot and object models from text, images, or their combinations~\citep{chen2026automatic,sontakke2026robotdesigngpt,wang2025kinematify}, while morphology--control co-design couples embodiment optimization with downstream control performance~\citep{qiu2024robomorph,fang2025robomore}.

\textbf{Policy Learning Guided by Reward and Motion.}
Multimodal large language models (MLLMs) are increasingly used to reduce manual specification in robot learning. One line of work synthesizes reward functions from textual task descriptions~\citep{kwon2023reward,yu2023language,xie2024text2reward,ma2024eureka}. Subsequent methods refine rewards using human feedback, training statistics, or execution feedback~\citep{hazra2025revolve,li2024auto,sun2025large,gao2026rf}. A complementary line uses reference motion to specify desired behavior. Classical motion-imitation and motion-prior methods enable complex locomotion and whole-body control~\citep{peng2018deepmimic,peng2021amp,escontrela2022adversarial,peng2022ase}, and recent systems scale this paradigm to large motion datasets and expressive humanoid skills~\citep{luo2023perpetual,fu2024humanplus,he2024omnih2o,liao2026beyondmimic}. More recently, MimicAgent uses MLLMs to generate task-specific reference motions for downstream RL~\citep{nayak2026mimicagentquadrupedskillsprompttotrajectory}, but without policy feedback.

Existing approaches typically improve either the learning objective or the reference behavior in isolation: reward-generation methods refine \emph{how} a policy is optimized, while motion-generation methods refine \emph{what} behavior the policy is asked to reproduce. In contrast, our framework uses policy rollouts to diagnose whether failure originates from the reference motion, the reward specification, or both, and revises the corresponding component accordingly. This creates a closed motion--policy loop in which execution feedback jointly shapes the behavioral target and its optimization guidance.

\section{Embodiment Creation}
\vspace{-0.5em}

Given a reference image and a set of task descriptions, we use GPT-6~\cite{gpt6} to generate a robot embodiment. The reference image specifies the desired coarse morphology, while the task descriptions provide functional cues about the motions the robot should support. We additionally provide CAD assets and specifications for candidate actuators, together with official assembly examples. In practice, we find that such assembly examples are important: they help the model infer how motors, links, fasteners, and supporting structures should be connected, rather than merely placing components in visually plausible configurations.
The generated embodiment $\mathcal{D}$ contains three consistent representations: (1) editable mechanical parts in STEP format for inspection and assembly, (2) an STL representation for fabrication, and (3) a URDF model for simulation and control. 

\vspace{-0.5em}
\subsection{Pre-Fabrication Feasibility Metrics}

Fabricating every MLLM-generated design is prohibitively expensive and provides feedback only after substantial time and material have been invested. We therefore introduce a set of computational checks that evaluate whether a generated embodiment is likely to be manufacturable, assemblable, and capable of supporting the intended behaviors before fabrication.

We group the checks into three categories according to the type of evidence they require:

\textbf{1) Design and Assembly} checks whether parts can be manufactured and fitted together. It verifies valid single-solid CAD parts of positive volume, watertight meshes positioned on the print bed, and matching volumes in source and exported models. It checks whether each part fits the printer's build volume in its printing orientation. 
At mating interfaces, it examines mounting holes, tool openings, and support material; worst-case tolerance checks verify that clearance remains under dimensional variation.
Assembly checks also examine insertion and tool-access paths, all fastener interfaces and traceable purchased-part specifications, and unwanted intersections in the complete assembly, including fasteners and cable envelopes.

\textbf{2) Mechanical and Actuation Feasibility} assess mechanical margins under the frozen material, load, and routing conventions. Local structural screening compares specified sections under recorded loads with an assumed allowable stress; load-path checks cover load-bearing parts and connections under defined load cases. Mass checks reconcile material estimates, link masses, and CAD and simulation totals. Fastener length chains compare selected screws with measured grip and nominal interface depths. Trajectory-limit margins check actual joint angles against existing limits; cable checks examine full-range clearance, bending, and twisting of the fixed routing candidate. Structural outcomes are conditional calculations, not measured strength; routing failures concern this candidate alone.

\textbf{3) Motion and Task Execution} evaluate whether the complete embodiment can realize the behaviors specified by the user. For each requested task, we construct a task-relevant reference motion by the agent and evaluate the robot along the corresponding trajectory. Checks include consistency of the simulated joint and inertial model, collision-free motion in the original CAD geometry, tracking error in simulation, actuator torque and saturation, joint-limit violations, and clearance margins under tracking error and fabrication tolerance.

Appendix~\ref{app:rg2-catalog} provides each check's criteria and implementation details.

\vspace{-0.5em}
\subsection{Evaluation}

We evaluate one gorilla embodiment generated by each of GPT-6 Astra, GPT-5.6 Sol, GPT-5.6 Terra and GPT-5.5. All models receive the same morphology reference, task descriptions, actuator candidates, and assembly examples. We evaluate each design using the proposed pre-fabrication metrics and the same two task motions: \emph{airborne forward swinging} and \emph{raised-hand waving}.

\begin{figure}[!htbp]
\centering
\includegraphics[width=0.8\linewidth]{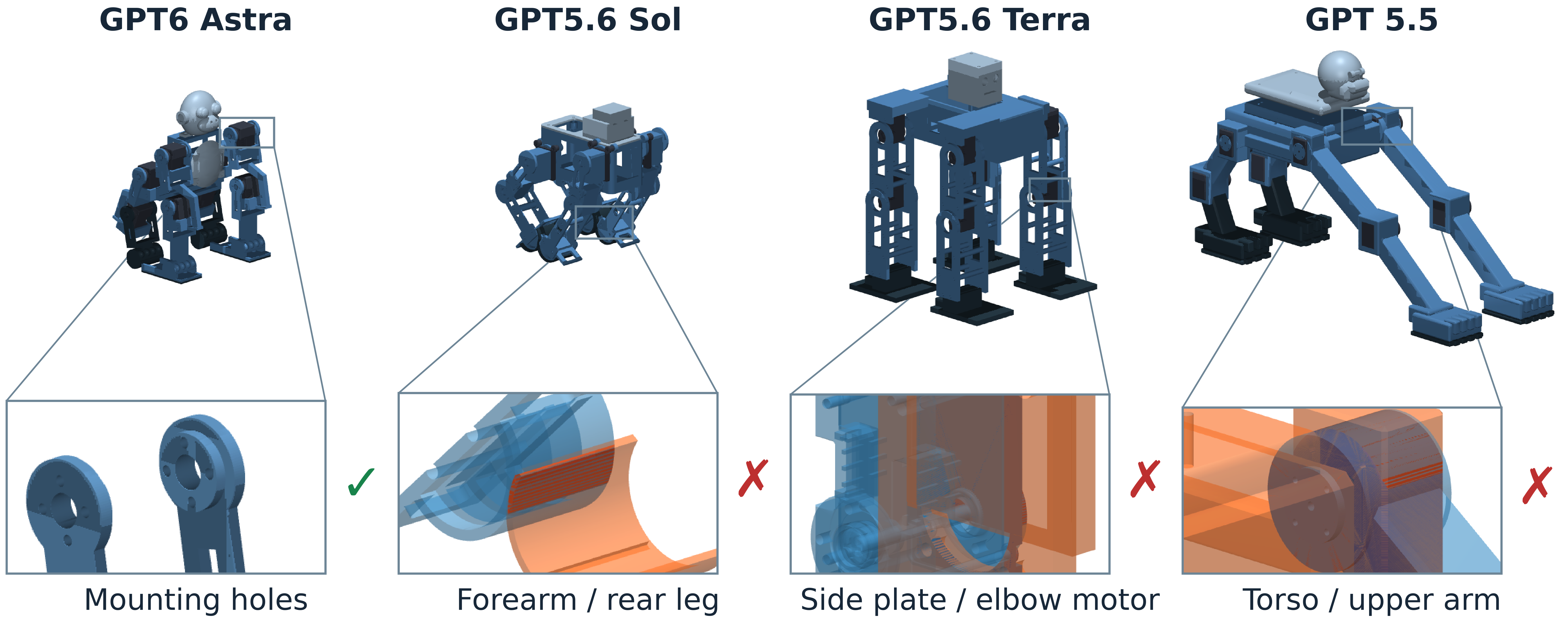}
\caption{Generated robots (top) and zoomed details  (bottom). GPT6-Astra shows the correct mounting-hole layout. The other views expose intersections between the forearm and rear leg in GPT5.6 Sol, the torso and upper arm in GPT~5.5, and the sideplate and elbow motor in GPT5.6 Terra. Surrounding parts are hidden or made transparent to reveal the interfaces.}
\label{fig:rg2-diagnostic}
\end{figure}

This comparison allows us to study not only whether a model produces a visually plausible robot, but also where its design fails mechanically. In Fig.~\ref{fig:rg2-diagnostic}, we observe qualitatively different failure modes across models, including missing mounting features, self-intersections between links, and insufficient clearance around actuators. Table~\ref{tab:robotgen-category-counts} reports the number of checks passed by each model. GPT6-Astra shows substantially stronger mechanical understanding than the other GPT variants. We then manually correct the remaining design issues and fabricate the embodiment via 3D printing.

\begin{table}[!htbp]
\centering
\setlength{\tabcolsep}{4pt}
\caption{
Pre-fabrication feasibility of robot embodiments generated by GPT variants.
Each entry $x/y$ denotes the number of passed checks among the $y$ checks in that category.
Appendix~\ref{app:rg2-catalog} provides the complete check definitions and per-design verdicts.
}
\label{tab:robotgen-category-counts}
\label{tab:rg2-pilot}
\begin{tabularx}{\linewidth}{@{}Xcccc@{}}
\toprule
Check category
& GPT6-Astra
& GPT5.6-Sol
& GPT5.6-Terra
& GPT 5.5 \\
\midrule
Design and Assembly
& 6/10 & 4/10 & 4/10 & 3/10 \\
Mechanical and Actuation Feasibility
& 4/6 & 0/6 & 1/6 & 0/6 \\
Motion and Task Execution
& 5/7 & 1/7 & 1/7 & 1/7 \\
\bottomrule
\end{tabularx}
\end{table}

\section{\method}
\label{sec:method}

\begin{figure}[!tbp]
  \centering
  \includegraphics[width=0.8\linewidth]{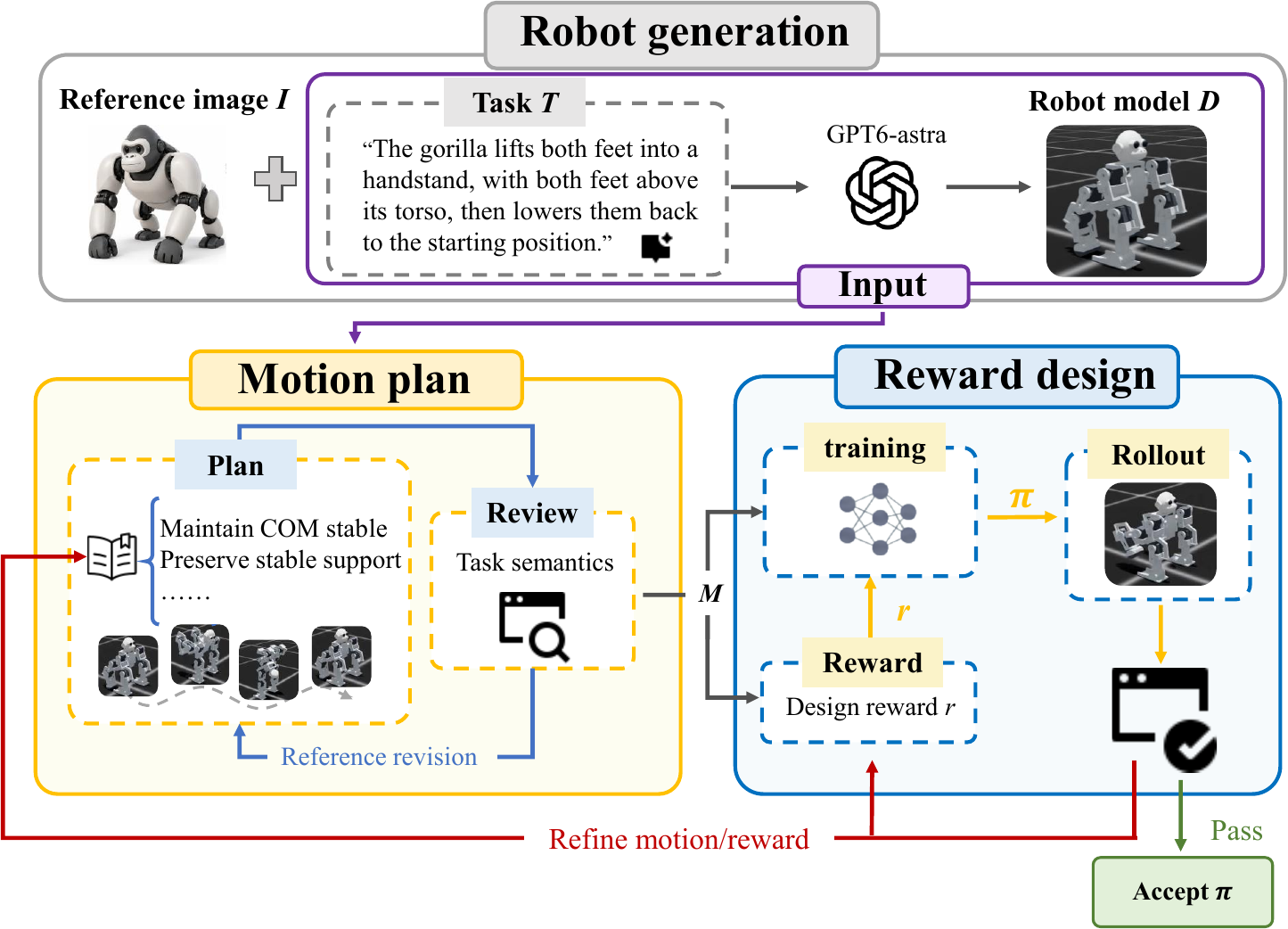}
  \caption{\textbf{Overview of {\method}: motion-guided, execution-grounded policy learning.}
Given a robot embodiment and task, the agent generates and validates a reference motion, configures the learning objective, and trains a policy. Execution feedback then refines the motion, reward, or both while keeping the embodiment and task fixed.}
  \label{fig:method-overview}
\end{figure}

We propose \textbf{{\method}} (Figure~\ref{fig:method-overview}), an agentic framework that learns executable robot behaviors from a given embodiment $D$ and task specification $T$. {\method} generates a reference motion $M$, configures a task-specific reward $r$, and trains a policy $\pi$, then uses policy execution to refine $M$, $r$, or both while keeping $D$ and $T$ fixed.

\subsection{Reference Motion Generation and Policy Learning}
\label{sec:reference-and-learning}

Given the robot model $D$ and task specification $T$, the agent instantiates the requested behavior through a robot-specific procedural motion specification. Rather than directly predicting joint trajectories, the agent specifies a root-pose trajectory $\mathbf{B}(t)$, end-effector targets $\mathbf{E}(t)$, and explicit trajectories $\mathbf{q}_{\mathcal K}(t)$ for selected task-relevant joints $\mathcal K$. Together, these components specify the whole-body motion, limb targets, and task-specific joint behavior.

The limb joints $\mathcal I$ that are not explicitly prescribed are then solved through inverse kinematics at each frame:
\begin{equation}
\mathbf q_{\mathcal I}^{*}(t)
=
\arg\min_{\mathbf q_{\mathcal I}}
\left\|
\mathbf e\!\left(
\mathbf q_{\mathcal I};
\mathbf B(t),\mathbf E(t),\mathbf q_{\mathcal K}(t)
\right)
\right\|^{2},
\qquad
\mathbf q_{\min,\mathcal I}
\leq
\mathbf q_{\mathcal I}
\leq
\mathbf q_{\max,\mathcal I},
\end{equation}
where $\mathcal I$ denotes the limb joints being optimized, $\mathbf B(t)$ is the prescribed root pose, $\mathbf E(t)$ denotes the target end-effector states, and $\mathbf q_{\mathcal K}(t)$ is held fixed during IK. The residual $\mathbf e$ measures the error between the resulting and target end-effector states. Combining the prescribed and IK-solved joint trajectories over all frames yields the joint-position trajectory $\mathbf Q\in\mathbb{R}^{N\times J}$, with joint velocities $\dot{\mathbf Q}\in\mathbb{R}^{N\times J}$ obtained by temporal differentiation. Forward kinematics further provides the corresponding world-frame body states, including body positions, orientations, linear velocities, and angular velocities. Together, these trajectories form a candidate reference motion $\widetilde M$.

The complete trajectory is checked for joint-position and velocity limits, self-collision, and consistency with the requested behavior, temporal structure, and style. Violations trigger revision of the procedural specification and re-solving. The accepted motion is denoted by $M$.

Using $T$ and $M$, the agent configures an initial reward $r$ from a DeepMimic\citep{peng2018deepmimic}-style reference-tracking library of composable tracking terms, for example, joint-pose, joint-velocity, and body-motion tracking terms, and trains a policy $\pi$ through reinforcement learning. Here, $M$ specifies a concrete realization of the behavior, while $r$ provides the optimization guidance for learning it. Together, they form the initial learning guidance, whose effectiveness is subsequently evaluated through policy rollouts.

\subsection{Execution-Grounded Motion and Reward Refinement}
\label{sec:execution-feedback}

A semantically valid and kinematically feasible reference does not necessarily provide effective learning guidance, as valid realizations can differ substantially in posture, timing, support, whole-body coordination, and learnability. We therefore evaluate policies through simulation rollouts for task-semantic compliance and execution quality, including motion quality, support, and balance.

We denote the initial motion, reward, and policy by $M_0=M$, $r_0=r$, and $\pi_0=\pi$. At refinement round $k$, $M_k$ and $r_k$ are used to train $\pi_k$. If the trial budget is exhausted, $\pi_k$ is returned without further review; otherwise, its rollout is evaluated against the same criteria. If successful, $\pi_k$ is returned; otherwise, the agent revises $M_k$, $r_k$, or both to obtain $(M_{k+1},r_{k+1})$.

\subsection{Learnability-Aware Execution Diagnosis and Revision}
\label{sec:refinement-guidance}

For each rollout that enters refinement, the agent localizes the failure across four levels: the intended task, the reference motion $M_k$ and body trajectory, the policy command $\mathbf q_{\mathrm{cmd}}(t)$, and the realized motion $\mathbf q_{\mathrm{actual}}(t)$ together with contact states. This helps distinguish limitations of the current motion realization from failures introduced during policy learning and execution. Tracking discrepancies are treated as diagnostic evidence rather than the evaluation objective.

The agent diagnoses whether the failure stems from $M_k$, considering motion structure, geometric and joint margins, timing, support transitions, and whole-body coordination, or from the optimization guidance in $r_k$. Motion-related failures lead to revisions of posture, trajectory, timing, support coordination, or phase structure while preserving task semantics and successful behavior; reward-related failures lead to adding, removing, or reweighting reward terms. Both may be revised when their effects interact. Each revision specifies the suspected cause, proposed change, behavior to preserve, and expected evidence in the next rollout. Section~\ref{sec:exp-guidance} evaluates this structured procedure.

\section{Experiments}
\label{sec:experiments}

\begin{table}[tbp]
  \caption{User-study ratings across four robot embodiments ($n=57$; $\uparrow$). {\method} achieves the highest overall rating and the highest rating on three of four embodiments.}
  \label{tab:user-study}
  \centering
  \begin{tabular*}{\linewidth}{@{\extracolsep{\fill}}lrrrrr@{}}
    \toprule
    Method & Go2 & MicroDuck & Ostrich & Gorilla & Overall \\
    \midrule
    Eureka-style    & 2.79 & 2.01 & 2.00 & 1.59 & 2.10 \\
    One-shot Motion & 3.93 & 3.30 & \textbf{4.04} & 3.14 & 3.60 \\
    \textbf{{\method} (Ours)}   & \textbf{4.10} & \textbf{4.03} & 3.99 & \textbf{3.77} & \textbf{3.97} \\
    \bottomrule
  \end{tabular*}
\end{table}

This section addresses three central questions:
\textbf{Q1: Does execution-grounded motion-reward refinement improve policy learning?}
\textbf{Q2: What does keeping motion in the loop change?}
\textbf{Q3: Can GPT recover the refinement strategy without explicit guidance?}
Together, these experiments evaluate the effectiveness of execution-grounded motion--reward refinement, examine how keeping motion in the loop allows execution feedback to reshape the concrete realization of a task toward more learnable behaviors, and test whether the GPT agent can independently recover and execute the proposed refinement strategy.

\subsection{Experimental Setup}
\label{sec:exp-setup}

\begin{table}[tbp]
  \caption{Reference-independent execution-quality metrics ($\downarrow$).
{\method} achieves the lowest mean value for every metric.}
  \label{tab:motion-quality}
  \centering
  \begin{tabular*}{\linewidth}{@{\extracolsep{\fill}}lrrrr@{}}
    \toprule
    Method
      & Yaw Err. (rad)
      & Foot-slip (m/s)
      & Action Var. (rad$^2$)
      & Joint Acc.  (rad/s$^2$) \\
    \midrule
    Eureka-style
      & 0.0345
      & 0.121
      & 0.0830
      & 114 \\
    One-shot Motion
      & 0.0584
      & 0.0716
      & 0.00665
      & 31.7 \\
    \textbf{{\method} (Ours)}
      & \textbf{0.00836}
      & \textbf{0.0131}
      & \textbf{0.00290}
      & \textbf{22.0} \\
    \bottomrule
  \end{tabular*}
\end{table}

\begin{figure}[tbp]
  \centering
  \includegraphics[width=0.9\linewidth]{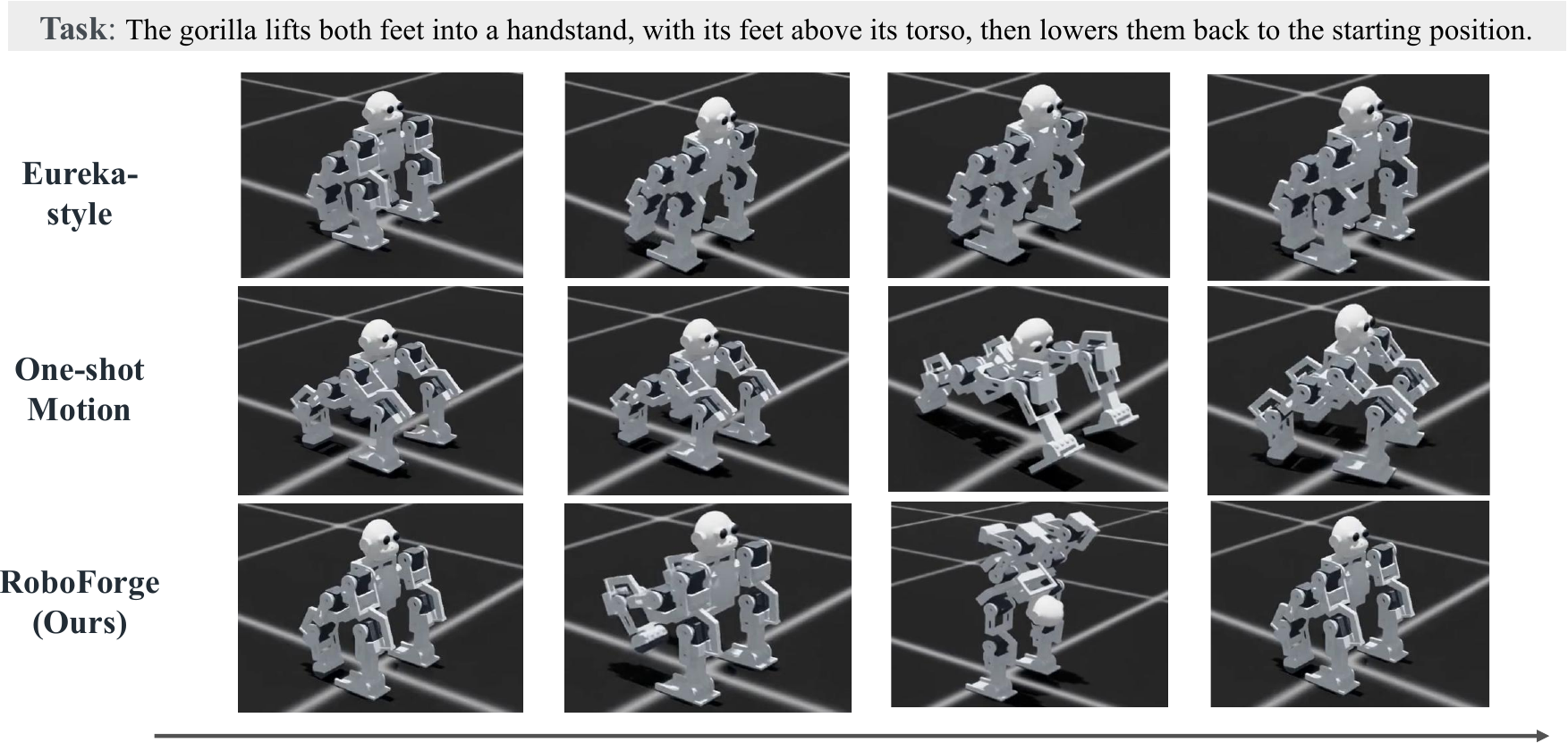}
  \caption{Qualitative comparison of learned policy executions on the gorilla handstand task. Eureka-style fails after an initial rear-leg lift, while One-shot loses balance with outward leg rotation. {\method} folds the legs inward before inversion, enabling a balanced transition and successful completion.}
  \label{fig:q1}
\end{figure}

\paragraph{Tasks and robots.}
We evaluate four embodiments: Unitree Go2, MicroDuck, and GPT-generated Ostrich and Gorilla, with the Gorilla additionally fabricated and evaluated on hardware. We use five natural-language tasks for each of the first three robots and four for Gorilla, yielding 19 tasks in total. The tasks span locomotion, stylized behaviors, and whole-body skills with diverse coordination and contact patterns, while admitting multiple realizations consistent with the same task semantics. Robot and task details are provided in Appendix~\ref{app:robots} and~\ref{app:tasks}, respectively.

\paragraph{Baselines.}
We compare \textbf{Eureka-style} and \textbf{One-shot Motion} with \textbf{{\method}}. Eureka-style, inspired by Eureka~\citep{ma2024eureka}, refines rewards from policy execution without reference motions. One-shot Motion follows a MimicAgent-style approach~\citep{nayak2026mimicagentquadrupedskillsprompttotrajectory}, using the same validated motion-generation pipeline as {\method} but fixing the reference once policy learning begins. {\method} instead refines the motion, reward, or both from execution feedback, with at most three training trials per task. All methods share the same PPO architecture, simulator configuration, and GPT-6 Astra agent~\citep{gpt6}. Further details are provided in Appendix~\ref{app:environment} and~\ref{app:implementation}.

\paragraph{Metrics.}
We evaluate final policy behavior using metrics independent of each method's reference motion and reward formulation, since neither reference-tracking errors nor optimized reward values are directly comparable across methods. Task-level behavior is assessed through a visual \textbf{user study}, where raters score compliance with the natural-language task and execution quality.
We also report four reference-independent quantitative metrics. For walking tasks, which require straight-line locomotion without a target speed, \textbf{Yaw Err.} measures directional compliance and \textbf{Foot-slip} characterizes contact stability. Across all tasks, Action Variation (\textbf{Action Var.}) and \textbf{Joint Acc.} measure smoothness at the command and realized-motion levels, respectively. Detailed definitions are provided in Appendix~\ref{app:metrics}.

\subsection{Q1: Does Execution-Grounded Motion-Reward Refinement Improve Policy Learning?}
\label{sec:exp-main}

\begin{table}[tbp]
  \caption{Comparison with unguided GPT. GPT spontaneously introduces and refines reference motions, while our structured pipeline achieves higher user ratings and better execution metrics. $^{*}$Motion generation is prescribed in our pipeline.}
  \label{tab:gpt-ours}

  \centering
  \setlength{\tabcolsep}{2.5pt}

  \begin{tabular*}{\linewidth}{@{\extracolsep{\fill}}lrrrrrrr@{}}
    \toprule
    & \multicolumn{2}{c}{Agent Behavior}
    & \multicolumn{5}{c}{Final Policy Quality} \\
    \cmidrule(lr){2-3}
    \cmidrule(l){4-8}
    Method
      & \shortstack{Motion\\Gen.}
      & \shortstack{Motion\\Refine}
      & \shortstack{User\\Rating $\uparrow$}
      & \shortstack{Yaw Err.\\$\downarrow$}
      & \shortstack{Foot-slip\\$\downarrow$}
      & \shortstack{Action Var.\\$\downarrow$}
      & \shortstack{Joint Acc.\\$\downarrow$} \\
    \midrule
    GPT
      & 16/19
      & 7/19
      & 3.61
      & 0.0172
      & 0.0419
      & 0.0179
      & 29.2 \\
    \textbf{{\method} (Ours)}
      & 19/19$^{*}$
      & 11/19
      & \textbf{3.97}
      & \textbf{0.00836}
      & \textbf{0.0131}
      & \textbf{0.00290}
      & \textbf{22.0} \\
    \bottomrule
  \end{tabular*}
\end{table}

Table~\ref{tab:user-study} compares {\method} with two complementary alternatives. Eureka-style retains reward refinement in the execution loop but provides no explicit motion guidance; {\method} improves the overall rating from 2.10 to 3.97, indicating the value of adding structured motion guidance. One-shot Motion provides such guidance but fixes the accepted reference once policy learning begins. Relative to this baseline, {\method} improves the overall score from 3.60 to 3.97 and achieves the highest rating on three of four embodiments, with only a slight decrease on Ostrich. Since One-shot Motion corresponds to the first policy-training trial of our pipeline, this comparison shows the benefit of keeping the initially generated learning guidance revisable after observing policy execution. Together, the two comparisons highlight complementary roles: motion provides an explicit behavioral realization, while execution-grounded refinement allows that realization and its learning objective to adapt when execution exposes their limitations.

The user-study results are supported by the reference-independent quantitative metrics in Table~\ref{tab:motion-quality}, where our method performs best on all four measures. Lower Yaw Err. and Foot-slip indicate more accurate directional control and cleaner support contacts, while lower Action Var. and Joint Acc. indicate smoother commands and realized motion. These results show that the user-rated improvements are accompanied by measurable gains in execution quality.

Figure~\ref{fig:q1} illustrates this process on the Gorilla handstand task. Eureka-style briefly lifts the rear legs but fails to complete the motion, while One-shot Motion rotates the lower legs outward and rapidly loses balance. Our method instead revises the realization by folding the lower legs inward before rotating the body upward, producing a more compact transition that completes the handstand. This comparison illustrates a key advantage of execution-grounded motion--reward refinement: rather than committing to an initially generated realization, the learning guidance can adapt after observing policy failure, allowing the resulting policy to adopt a realization that is better aligned with successful execution.

\subsection{Q2: What Does Keeping Motion in the Loop Change?}
\label{sec:exp-motion-loop}
\begin{figure}[tbp]
  \centering
  \includegraphics[width=0.7\linewidth]{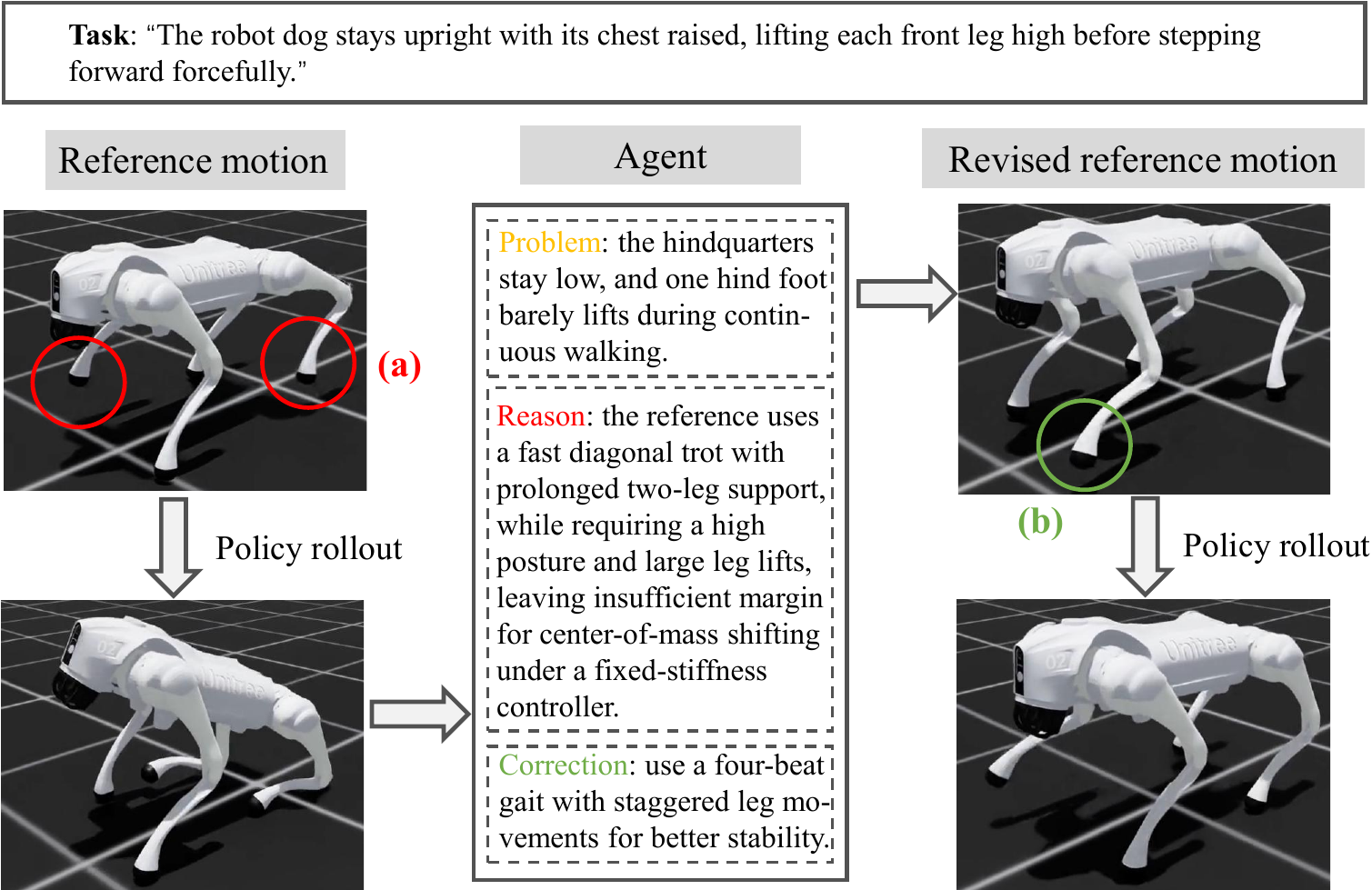}
  \caption{Motion refinement preserves task semantics while improving gait realization. The initial gait has two airborne feet (a), whereas the refined gait uses single-foot swing (b) with coordinated body shift, yielding a more stable four-beat gait.}
  \label{fig:q2}
\end{figure}

Keeping motion in the loop allows the agent to revise \emph{how} a task is realized while preserving \emph{what} it requires. As shown in Figure~\ref{fig:q2}, both references satisfy the requested high-posture, high-stepping walk, but the initial two-beat diagonal trot is difficult to reproduce, with rollouts showing rear-leg support collapse and insufficient hind-foot clearance. The agent attributes this difficulty to limited center-of-mass transfer under two-leg support, high posture, and large leg excursions, and revises the motion to a four-beat gait with staggered limb swings and coordinated body shifts. This preserves task semantics while providing a more learnable realization, showing that execution feedback can revise a valid but difficult-to-realize reference rather than only the learning objective.

Across the 19 evaluated tasks, the framework revises the generated motion in 11 tasks, with 15 revisions in total. This prevalence shows that execution frequently exposes limitations missed during motion generation, making refinement a recurring step in bridging generation and execution.

\subsection{Q3: Can GPT Discover the Refinement Strategy Without Explicit Guidance?}
\label{sec:exp-guidance}

To test whether the refinement strategy emerges without explicit guidance, we remove the structured procedure and ask GPT to solve each task through reinforcement learning using only the shared policy architecture and training interface. Without instructions to generate or revise reference motions, GPT introduces motion guidance in 16 of 19 tasks and refines it in 7. This shows that motion-guided learning and execution-driven motion revision can emerge independently in the unguided GPT agent.

Discovering the strategy does not guarantee effective execution. Table~\ref{tab:gpt-ours} shows that our structured pipeline achieves higher user ratings and better results on all four reference-independent metrics, with especially large reductions in foot slip and action variation. This indicates that explicit guidance helps translate execution failures into more effective motion and reward revisions, rather than merely prompting refinement itself.

\begin{figure}[tbp]
  \centering
  \includegraphics[width=0.9\linewidth]{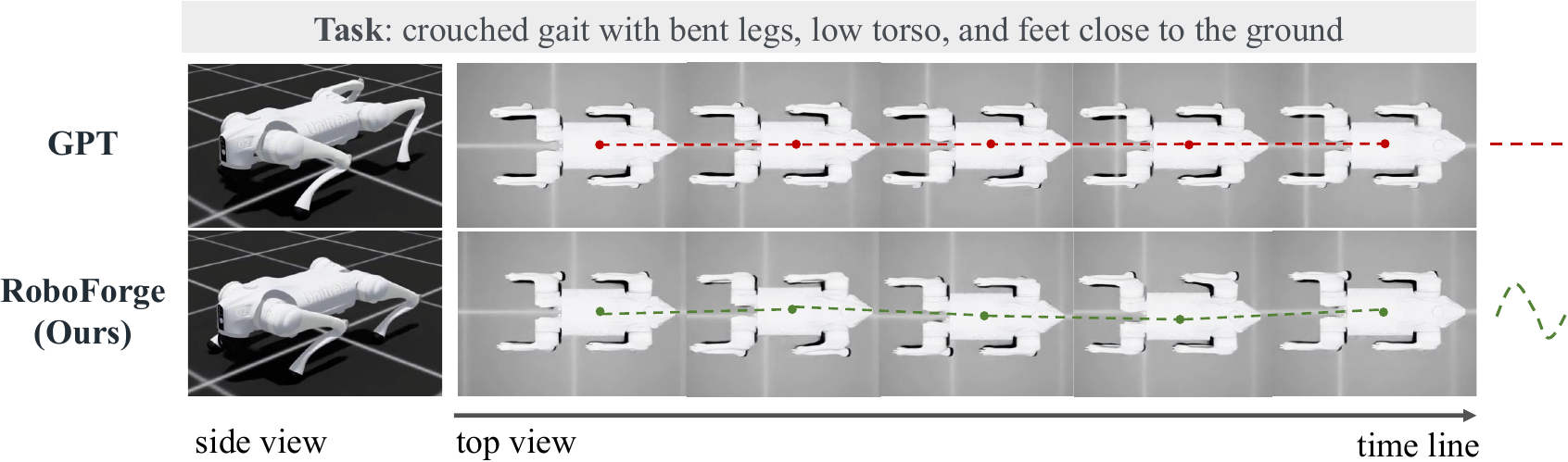}
  \caption{Refined motions from GPT and {\method}. Both increase swing-foot clearance, but differ in whole-body coordination. CoM trajectories show that GPT remains nearly straight, whereas {\method} shifts laterally toward the supporting legs for coordinated weight transfer.}
  \label{fig:q3}
\end{figure}

Figure~\ref{fig:q3} illustrates this distinction. Both approaches observe foot sliding and increase swing-foot clearance, but differ in how they revise the motion. Unguided GPT primarily corrects the visible gait deficiency, whereas our framework also revises body--support coordination by shifting the body toward the stance legs during swing while preserving their support trajectories in world coordinates. The resulting motion therefore improves not only gait appearance, but also the whole-body realization presented for policy learning. Further results are provided in Appendix~\ref{app:unguided-gpt}.

\section{Real-World Demonstration}
\label{sec:real-world}

\begin{figure}[tbp]
  \centering
  \IfFileExists{figures/real_robot.jpg}{%
    \includegraphics[width=0.7\linewidth]{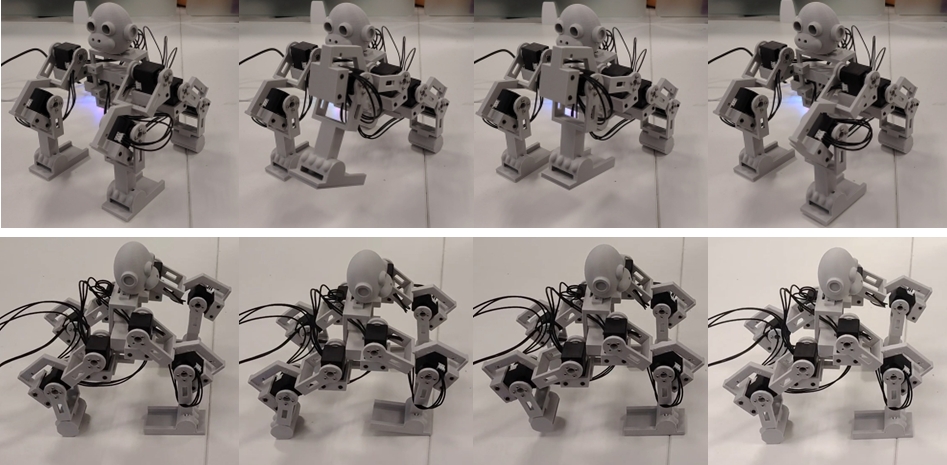}%
  }{%
    \figureplaceholder{4.0cm}{Hardware photographs to be supplied.
    Top row: hand waving.
    Bottom row: walking.}%
  }
  \caption{Physical realization of the generated Gorilla and its learned behaviors.
  Top: hand waving. Bottom: walking. Frames are ordered from left to right.}
  \label{fig:gorilla-real-world}
\end{figure}

We validate {\method} on hardware using the agent-generated 8-DoF Gorilla, which we 3D-print and assemble (Appendices~\ref{app:gorilla-kinematics} and~\ref{app:hardware_fabrication}). We directly deploy the simulation-trained walking and hand-waving policies: walking requires repeated hand--foot support transfer, while hand waving maintains support through the remaining limbs (Figure~\ref{fig:gorilla-real-world}). These demonstrations extend the behavior-learning results from simulation to physical execution on the fabricated robot.

\section{Conclusion}
\label{sec:conclusion}

We studied agentic robot creation from two complementary aspects: embodiment and behavior creation. For embodiment creation, our computational metrics expose mechanically relevant failures in MLLM-generated designs before fabrication. For behavior creation, we presented \textbf{{\method}}, which learns executable policies through motion generation and execution-grounded refinement. Across four embodiments, {\method} improves task-level behavior and execution quality, while direct deployment on the fabricated Gorilla demonstrates learned behaviors on hardware. Together, these results highlight the value of physical verification for generated embodiments and adaptive motion guidance for behavior learning.

\subsection*{AI Use Statement}
Generative AI tools were used during this work to assist with research ideation and execution, to draft portions of the manuscript, and to aid with language editing and polishing. AI systems also constitute part of the proposed research framework and experimental methodology, including robot generation and execution-grounded motion and reward refinement; these uses are described explicitly in the corresponding method and experimental sections. All research decisions, experimental design, implementation, analysis, verification of results, and final manuscript content were reviewed and approved by the authors.

\subsection*{Ethics Statement}
Our user study involved human participants. All participants were informed of the purpose and procedure of the study and provided informed consent before participation. Participation was voluntary, and participants could withdraw at any time. We did not collect personally identifiable or sensitive information, and all reported results were aggregated and anonymized. The study was conducted in accordance with our institution's requirements for research involving human participants.

\subsection*{Reproducibility Statement}
We provide the key implementation and evaluation details used in our experiments in the Appendix, including the policy-training setup, baseline configurations, metric definitions, and user-study protocol. Each refinement trial retrains the policy from scratch without reusing previous checkpoints. Additional experimental settings are reported where necessary to interpret and reproduce the presented results.

\bibliography{iclr2027_conference}
\bibliographystyle{iclr2027_conference}

\appendix
\clearpage
\section{Robot Embodiments}
\label{app:robots}
\label{app:supplement}
\subsection{Embodiment Overview}
\label{app:robot-overview}

\begin{figure}[!hbp]
  \centering

  \begin{minipage}[b]{0.48\linewidth}
    \centering
    \reflectbox{
      \includegraphics[
        height=0.23\textheight,
        width=0.95\linewidth,
        keepaspectratio
      ]{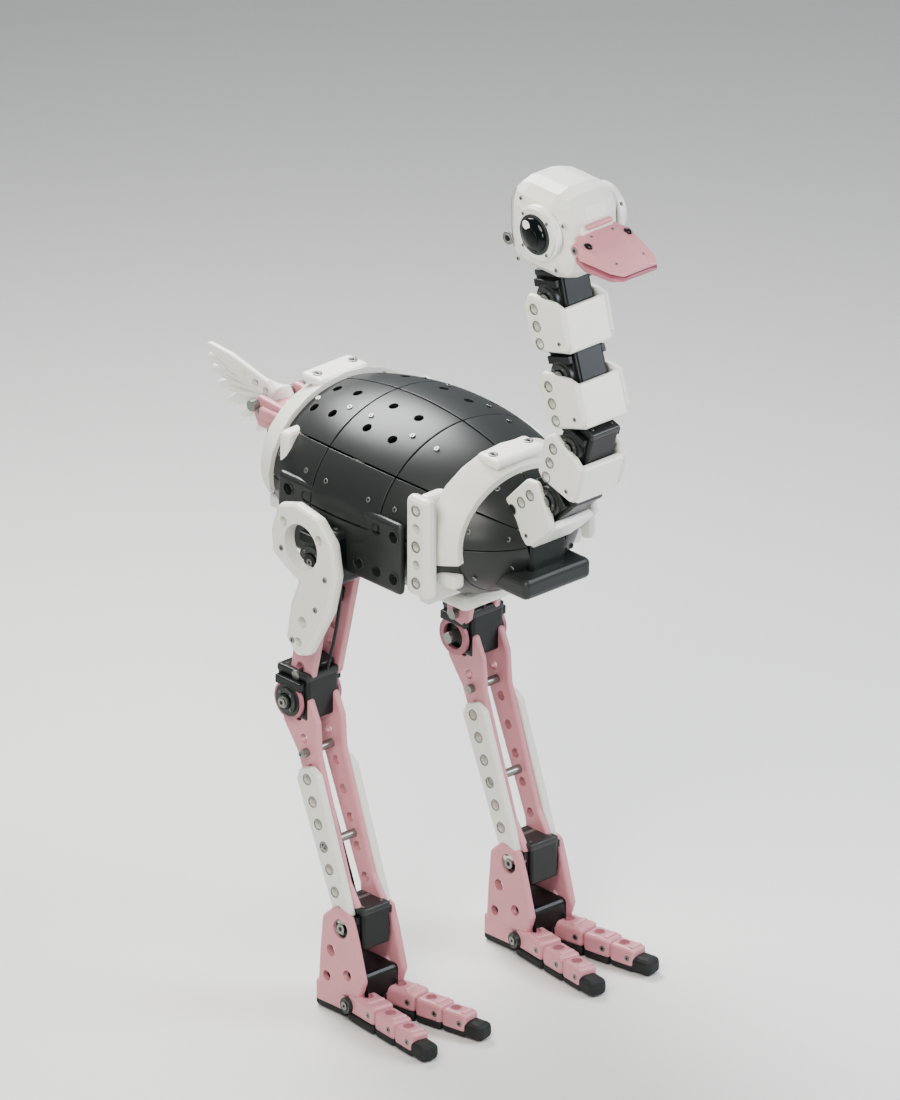}
    }
    \par\vspace{2pt}
     (a) Ostrich
  \end{minipage}
  \hfill
  \begin{minipage}[b]{0.48\linewidth}
    \centering
    \includegraphics[
      height=0.23\textheight,
      width=0.95\linewidth,
      keepaspectratio
    ]{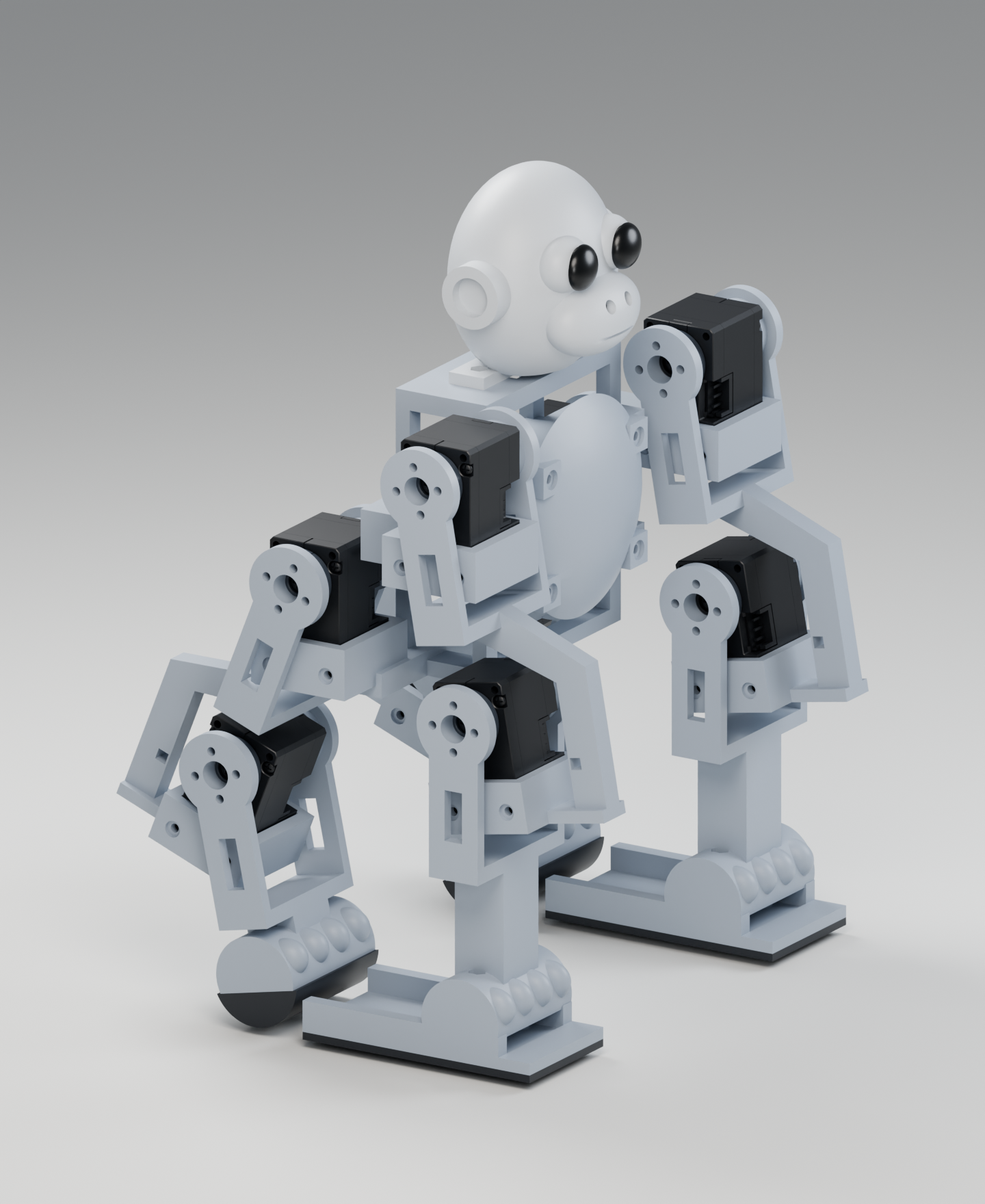}
    \par\vspace{2pt}
     (b) Gorilla
  \end{minipage}

  \caption{
    GPT-generated robot embodiments used in our experiments.
    (a) Ostrich.
    (b) Gorilla, whose design is additionally validated through physical fabrication and assembly.
  }
  \label{fig:app-robots}
\end{figure}

We evaluate our framework on four robot embodiments: the Unitree Go2, MicroDuck, and two robots generated by GPT-6 Astra, Ostrich and Gorilla. We particularly highlight Gorilla as a physically realized generated embodiment. Its construction explicitly accounts for the real motor model, self-collisions that arise during target-task execution, and manufacturability, enabling the resulting design to be directly fabricated and assembled. Figure~\ref{fig:app-robots} shows the GPT-generated Ostrich and Gorilla. Once an embodiment is constructed, its structure and joint definitions remain fixed throughout the subsequent motion--RL loop; refinement operates only on the learning guidance for that embodiment.

\subsection{Ostrich Kinematic Tree}
\label{app:ostrich-kinematics}

Ostrich has 15 actuated degrees of freedom, each driven by a single motor. Its kinematic tree consists of 18 links connected by 17 joints, including 15 revolute joints and two fixed joints. The torso (\texttt{base\_link}) forms the root, from which four branches extend to the neck and head, the left leg, the right leg, and the tail.

\paragraph{Neck and head: six motors.}
The neck and head form a six-DoF serial chain. Five pitch joints connect the torso to four successive neck links and then to the head mount, followed by a yaw joint connecting the head mount to the head. This branch therefore contains six links beyond the torso: four neck links, one head mount, and one head link.

\paragraph{Left and right legs: four motors each.}
Each leg branches from the torso through four actuated joints: hip roll, hip pitch, knee pitch, and ankle pitch. The corresponding link chain consists of the hip-roll link, thigh, shin, ankle, and foot. The foot is rigidly attached to the ankle through a fixed joint and therefore introduces no additional actuated degree of freedom. Together, the two legs account for eight motors.

\paragraph{Tail: one motor.}
A single pitch joint connects the torso directly to the tail link. Overall, the 15 motors are distributed as six across the neck and head, four on each leg, and one on the tail.

\subsection{Gorilla Kinematic Tree}
\label{app:gorilla-kinematics}

Gorilla has eight actuated degrees of freedom, each driven by a single motor. Its kinematic tree consists of nine links connected by eight revolute joints. The torso (\texttt{base}), which also incorporates the head, forms the root of four branches corresponding to the left arm, right arm, left leg, and right leg.

\paragraph{Left and right arms: two motors each.}
Each arm contains two pitch joints. A shoulder-pitch joint connects the torso to the upper arm, followed by an elbow-pitch joint connecting the upper arm to the lower arm. The hand is integrated into the lower-arm link and therefore does not introduce a separate wrist joint or actuated degree of freedom. Each arm consequently contains two links and two actuated joints.

\paragraph{Left and right legs: two motors each.}
Each leg similarly contains two pitch joints. A hip-pitch joint connects the torso to the upper leg, followed by a knee-pitch joint connecting the upper leg to the lower leg. The foot is integrated into the lower-leg link and therefore introduces no separate ankle joint or actuated degree of freedom. Each leg consequently contains two links and two actuated joints.

All eight actuated joints are pitch joints, yielding two motors per limb and eight motors in total.

\subsection{Gorilla Hardware Fabrication}
\label{app:hardware_fabrication}

\begin{figure}[tbp]
  \centering
  \includegraphics[width=0.75\linewidth]{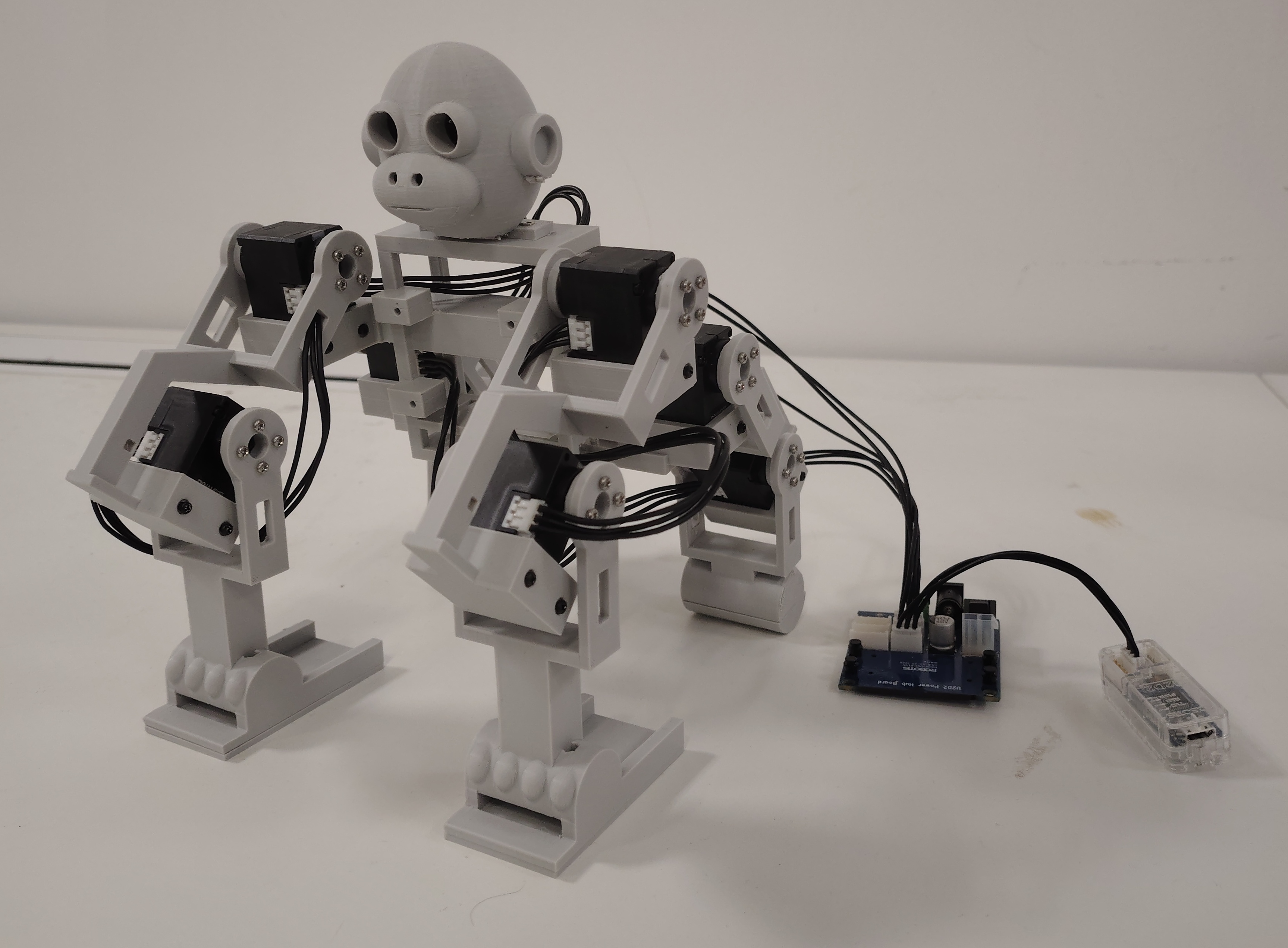}%
  \caption{Fabricated Gorilla embodiment.}
  \label{fig:gorilla_hardware}
\end{figure}

We fabricated the agent-generated Gorilla from its
verified CAD design. The structural components were
3D printed in PLA using a Bambu Lab X1E printer with a layer height of
0.2\,mm and an infill density of 15\%.
Figure~\ref{fig:gorilla_hardware} shows the resulting physical embodiment.

The electrical and actuation hardware comprises eight DYNAMIXEL
XL330-M288-T actuators, a U2D2 communication interface, a U2D2 Power Hub
Board Set, an external power supply, and a host PC. The U2D2 bridges the
PC's USB connection to the actuators' TTL serial bus, allowing the PC to
send joint commands and receive actuator feedback using DYNAMIXEL
Protocol~2.0. The Power Hub Board distributes power from the external supply
to the actuators.

\section{Simulation and Reinforcement Learning Setup}
\label{app:environment}

This section specifies the common simulation environment, actor architecture,
observation interface, and PPO configuration used as the reinforcement-learning
template. Unless otherwise noted, the settings below define the common
starting configuration, with task-specific settings specified by each method
when required. Task objectives and refinement procedures are described in
Appendix~\ref{app:implementation} and ~\ref{app:task-inventory}.

\subsection{Simulation and Control Settings}
\label{app:simulation-settings}

Experiments use Isaac Lab 2.1 with Isaac Sim 4.5. The scene contains a flat
ground plane at $z=0$, with static and dynamic friction coefficients of $1.0$
and restitution $0$. Each robot is controlled at 50 Hz across all four
methods, corresponding to one policy update every $0.02$ s. Joint-position
targets are held between policy updates while the simulator advances the
robot dynamics.

Table~\ref{tab:app-simulation} lists the physics timestep
$\Delta t_{\mathrm{sim}}$ and control decimation $d$, i.e., the number of
physics steps per policy update. These satisfy
$f_{\pi}=1/(\Delta t_{\mathrm{sim}}d)=50$ Hz. Go2, MicroDuck, and Ostrich use
a $0.005$ s physics timestep with $d=4$, whereas Gorilla uses a $0.002$ s
timestep with $d=10$.

\begin{table}[htbp]
\centering
\caption{Simulation timing shared across methods. All robots are controlled
at a policy frequency of 50 Hz.}
\label{tab:app-simulation}
\begin{tabular*}{\linewidth}{@{\extracolsep{\fill}}lrrr@{}}
\toprule
Robot & Physics step (s) & Decimation & Policy (Hz) \\
\midrule
Go2, MicroDuck, Ostrich & 0.005 & 4 & 50 \\
Gorilla & 0.002 & 10 & 50 \\
\bottomrule
\end{tabular*}
\end{table}

\subsection{Noise and Initialization}

All methods use the same nominal home configuration for each robot and are equally permitted to specify a task-dependent initial pose when appropriate for the target behavior.

By default, the motion-tracking configurations used by One-shot Motion and {\method} add no observation noise to base velocities, projected gravity, joint positions, joint velocities, or observation history. Joint-position reset perturbations and reset-velocity scaling are initialized to zero, contact-material parameters are fixed, and no center-of-mass offsets or external pushes are applied.

The GPT and Eureka-style templates likewise disable observation corruption by default. Task-specific implementations may specify reset states, episode lengths, termination conditions, and physical perturbations. The base environment uses a default episode length of $10$ s. 

\subsection{Actor Network and Observation Template}

All methods use a feedforward actor with three hidden layers of widths
$512$, $256$, and $128$. Each hidden layer uses an ELU activation, followed
by a linear output layer. Given an observation vector
$o_t\in\mathbb{R}^{d_o}$, the actor produces an action mean in
$\mathbb{R}^{d_a}$ through
\begin{equation}
d_o \longrightarrow 512;\mathrm{(ELU)}
\longrightarrow 256;\mathrm{(ELU)}
\longrightarrow 128;\mathrm{(ELU)}
\longrightarrow d_a.
\end{equation}

The base observation concatenates the velocity command
$\mathbf{c}_t=[v_x^{\mathrm{cmd}},v_y^{\mathrm{cmd}},\omega_z^{\mathrm{cmd}}]$,
body-frame projected gravity $\mathbf{g}_t^b$, joint positions relative to
the nominal configuration, joint velocities, and the previous raw action:
\begin{equation}
o_t =
[\mathbf{c}_t,\mathbf{g}_t^b,
\mathbf{q}_t-\mathbf{q}_0,\dot{\mathbf{q}}_t,\mathbf{a}_{t-1}],
\qquad
d_o=6+2J+d_a,
\end{equation}
where $J$ denotes the number of joints and $\mathbf{q}_0$ the nominal joint
configuration used by the task interface. Joint positions and velocities are
expressed in rad and rad/s, respectively, while the command specifies
body-frame planar velocity in m/s and yaw rate in rad/s. The base interface
uses the current observation without a history window. The actor observation
may be extended only with signals available to the policy at inference time,
while retaining consistent feature definitions, joint mappings, and temporal
interpretation between training and inference. Additional observations may be
used by training-only components, such as the critic or, where applicable, a
discriminator, but are not provided to the actor.

The PPO runner applies empirical observation normalization,
\begin{equation}
\hat{o}_t =
\frac{o_t-\boldsymbol{\mu}_o}
{\boldsymbol{\sigma}_o+10^{-2}},
\end{equation}
where the running mean and standard deviation are estimated during training.
Statistics stop updating after $10^8$ observed samples and are stored with the
policy for inference.

During training, the policy samples raw actions from a diagonal Gaussian whose
mean is produced by the actor. Each action dimension has a learnable,
state-independent standard deviation initialized to $1.0$. At inference, the
policy uses the action mean. Task-specific action mappings convert raw actions
into valid joint-position targets using the corresponding scaling, offsets,
and joint limits.

\subsection{PPO Optimization Defaults}
\label{app:ppo-settings}

The shared RL template uses RSL-RL PPO with $4{,}096$ parallel environments
and collects $24$ policy steps per environment before each optimization phase.
Each iteration therefore contains $98{,}304$ transitions and spans $0.48$ s
of simulated control per environment at 50 Hz. The rollout batch is divided
into four minibatches of $24{,}576$ transitions and optimized for five epochs,
resulting in $20$ optimizer updates per iteration.

\begin{table}[htbp]
\centering
\caption{Default PPO and rollout configuration of the shared RL template.}
\label{tab:app-ppo-defaults}
\begin{tabular*}{\linewidth}{@{\extracolsep{\fill}}ll@{}}
\toprule
Parameter & Default value \\
\midrule
Parallel environments & $4{,}096$ \\
Policy steps per environment per iteration & $24$ \\
Transitions per iteration & $98{,}304$ \\
Optimization epochs per iteration & $5$ \\
Minibatches per epoch & $4$ \\
Transitions per minibatch & $24{,}576$ \\
\midrule
Optimizer & Adam \\
Initial learning rate & $10^{-3}$ \\
Learning-rate schedule & Adaptive to KL divergence \\
Target KL divergence & $0.01$ \\
Discount factor $\gamma$ & $0.99$ \\
GAE parameter $\lambda$ & $0.95$ \\
PPO clipping parameter & $0.2$ \\
Entropy coefficient & $0.005$ \\
Value-loss coefficient & $1.0$ \\
Clipped value loss & Enabled \\
Maximum gradient norm & $1.0$ \\
\midrule
Advantage normalization & Full rollout batch \\
Observation normalization & Empirical running statistics \\
Initial action standard deviation & $1.0$ \\
\bottomrule
\end{tabular*}
\end{table}

Advantages are computed using generalized advantage estimation and normalized
over the full rollout batch. The adaptive learning-rate schedule divides the
learning rate by $1.5$ when the mean KL divergence exceeds $0.02$ and
multiplies it by $1.5$ when the divergence lies in $(0,0.005)$, with the
learning rate bounded to $[10^{-5},10^{-2}]$.

The default random seed is $0$ and can be specified at launch. Checkpoints
are saved every $500$ iterations and once at the end of training. For GPT,
One-shot Motion, and {\method}, training uses a cumulative budget of up to $3{,}000$
iterations per trial; Eureka-style uses $2{,}250$ iterations per candidate.
Resumed runs restore the policy, optimizer, observation-normalization
statistics, and completed iteration count before continuing within the
remaining trial budget. Trial counts and per-task budgets are specified in
Appendix~\ref{app:method-budgets}.

\section{Task Diversity and Motion Realizations}
\label{app:tasks}

\subsection{Task Coverage}
\label{app:task-inventory}

We evaluate 19 natural-language tasks across four robot embodiments, including
seven walking tasks and twelve skills. The task set spans locomotion,
stylized behavior, and whole-body coordination, with substantial variation in
both the body parts involved and the required contact patterns.
Table~\ref{tab:app-task-inventory} lists the complete task inventory. Stylistic
requirements appear throughout both walking and skill tasks; for example,
locomotion may specify not only forward movement but also a recognizable
posture or manner of execution.

\begin{table}[htbp]
\centering
\caption{Task inventory across four embodiments: seven walking tasks and
twelve whole-body skills.}
\label{tab:app-task-inventory}
\begin{tabular*}{\linewidth}{@{\extracolsep{\fill}}p{0.13\linewidth}p{0.28\linewidth}p{0.40\linewidth}r@{}}
\toprule
Robot & Walking tasks & Whole-body skills & Total \\
\midrule
Go2 & High-step walk; low walk
    & Bow; look around; kick & 5 \\
\addlinespace
MicroDuck & Fast walk; waddling walk
          & Bow; look around; kick & 5 \\
\addlinespace
Ostrich & Happy walk; sad walk
        & Nod; stretch; squat and rise & 5 \\
\addlinespace
Gorilla & Alternating walk
        & Handstand; stretch; wave & 4 \\
\midrule
Total & 7 walking tasks & 12 skills & 19 \\
\bottomrule
\end{tabular*}
\end{table}

The required contact patterns range from posture changes under stationary
support to alternating stepping, transient single-foot support, and
hand-supported inversion. Successful execution therefore depends on
coordinating expressive motion with the limbs responsible for support and
balance. Task completion cannot in general be characterized by locomotion
progress alone, motivating evaluation at the level of the requested behavior
and its execution quality.

Importantly, the task instruction constrains the behavior without uniquely
determining its motion realization. It may specify a recognizable style,
action order, or required support phase, while leaving quantities such as
motion amplitude, timing, posture, and inter-limb coordination underdetermined.
This flexibility is what enables execution-grounded refinement: the
motion--RL loop can modify how a task is physically realized when the current
reference is difficult to learn or execute, while preserving the task
semantics specified by the user.

\subsection{Stylized Locomotion}
\label{app:walking-styles}

\begin{figure}[htbp]
  \centering
  \IfFileExists{figures/app_walk_ostrich.jpg}{%
    \includegraphics[width=\linewidth]{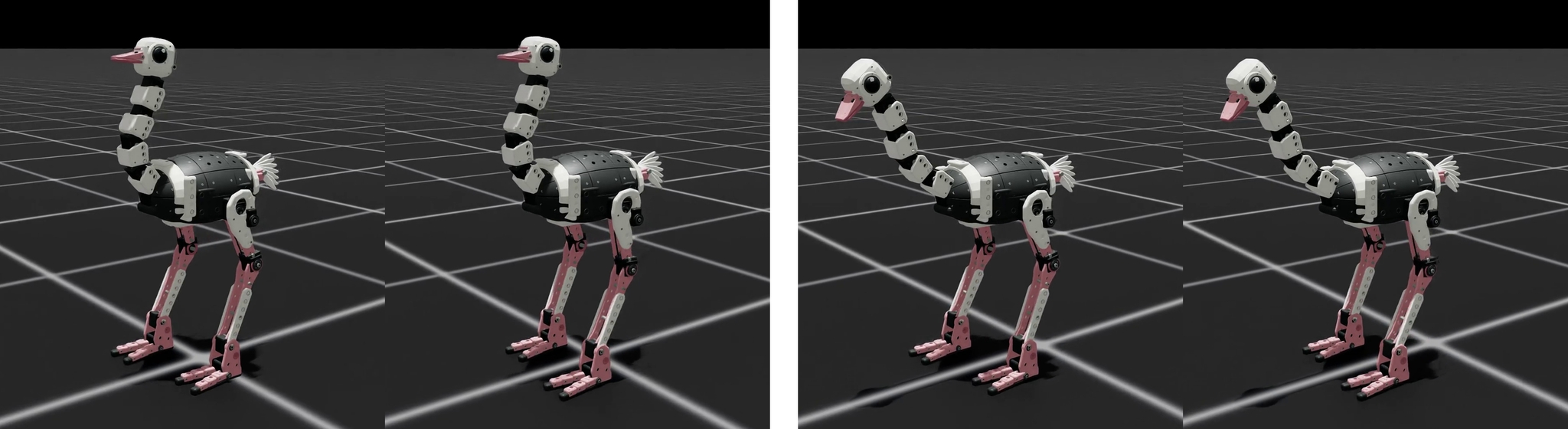}%
  }{%
    \figureplaceholder{2.7cm}{Ostrich walking styles.}%
  }
  \caption{Generated stylized locomotion motions for Ostrich.
  Left: \emph{happy walking}; right: \emph{sad walking}.
  Both motions are generated by our method and share the same
  forward-locomotion objective, while differing in head--neck
  posture and its coordination with the stepping pattern.}
  \label{fig:app-ostrich-walk-styles}
\end{figure}

Ostrich's two walking tasks illustrate how different motion realizations can share the same underlying locomotion objective (Figure~\ref{fig:app-ostrich-walk-styles}). As with all walking tasks in our evaluation, both require straight-line forward locomotion. The happy walk calls for an upright body, light alternating steps, and accompanying head motion, whereas the sad walk requires a lowered head while maintaining alternating forward progression. Their distinction therefore lies not in the locomotion direction, but in how head--neck posture and stepping are coordinated to express the requested style.

\subsection{Whole-Body Coordination}
\label{app:skill-tasks}

\begin{figure}[htbp]
\centering
\IfFileExists{figures/microduck_kick.jpg}{%
\includegraphics[width=\linewidth]{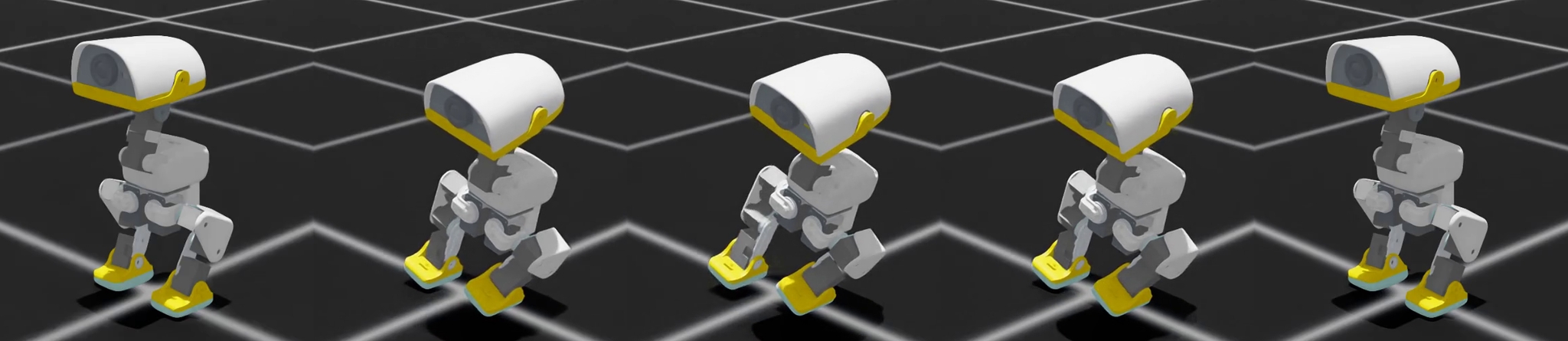}%
}{%
\figureplaceholder{2.7cm}{Figure to be supplied: MicroDuck kick sequence,
showing preparation, the kick under single-foot support, and recovery.}%
}
\caption{Generated motion for MicroDuck's kick task produced by our method. The sequence coordinates gaze, lateral body shift, the kicking leg, and the supporting leg across preparation, single-foot support, and recovery.}
\label{fig:app-microduck-kick}
\end{figure}

MicroDuck's kick illustrates the phased coordination required by a
whole-body skill (Figure~\ref{fig:app-microduck-kick}). The requested behavior
begins by looking downward and shifting the body toward one side, creating
support for lifting the opposite foot. The lifted foot then performs a short
forward kick before retracting as the robot returns to a stable two-foot
stance. The task semantics therefore include not only the kick itself, but
also the preparation, support transition, and recovery required to realize
the complete behavior.

\section{Method Implementation}
\label{app:implementation}

All agent roles across the compared methods use GPT-6 Astra with reasoning
effort set to \texttt{high}. The methods receive the same robot assets and
natural-language task specifications, while differing in the learning
scaffold they are given and in how policy execution is used to construct or
revise learning guidance.

\subsection{Unguided GPT and Our Motion--RL Loop}
\label{app:gpt-ours-implementation}

GPT and {\method} use the same training budget of up to three trials per task.
Each trial runs $4{,}096$ parallel environments and is limited to a cumulative
budget of $3{,}000$ PPO iterations; resumed training within a trial consumes
the remaining budget of that trial. Evaluation uses the terminal policy from
the last completed trial.

\paragraph{GPT.}
The unguided GPT baseline is provided with the shared actor template, the
robot and task specification, and general RL tools for constructing rewards,
training policies, and evaluating execution. It is not given a prescribed
procedure for reference-motion generation, motion refinement, or
learnability-aware diagnosis, nor is it provided with the reference-tracking
guidance used by {\method}. Within the same training budget, the agent determines
how to construct the learning objective and organize its training process.
No motion-based strategy is prohibited: the agent may generate, use, or
revise a reference motion if it chooses to do so. This setting examines
whether a general-purpose agent independently discovers and executes a
motion-based refinement strategy from task execution feedback.

\paragraph{{\method}.}
{\method} is provided with a DeepMimic-style reference-conditioned RL scaffold and
explicit guidance for generating, validating, and iteratively refining the
learning guidance. Before policy training, the generated reference motion is
checked for kinematic validity and task-semantic consistency. After training,
policy execution is evaluated against the requested behavior and used to
diagnose deficiencies in posture, timing, support transitions, and
whole-body coordination. The resulting feedback can revise the reference
motion, the reward, or both, while the robot embodiment and task semantics
remain fixed.

This comparison therefore tests whether the execution-grounded
motion--reward refinement strategy emerges autonomously from a general RL
interface or benefits from being explicitly instantiated as part of the
agentic learning procedure. The full refinement procedure is described in
the main Method section, while Appendix~\ref{app:unguided-gpt} analyzes the
strategies observed in the unguided GPT baseline.

\subsection{MimicAgent-style One-shot Motion}
\label{app:oneshot-implementation}

\paragraph{One-shot Motion. }Our One-shot Motion baseline follows the generate--validate--train structure
of MimicAgent~\citep{nayak2026mimicagentquadrupedskillsprompttotrajectory}. A reference motion is first
generated and validated, and the resulting motion is then used to guide
imitation-based reinforcement learning. MimicAgent itself allows trajectory
generation to iterate in response to task-agnostic validity tests before
policy training. Accordingly, ``one-shot'' refers to the absence of execution-driven motion refinement after policy learning begins, rather than to generating the reference motion only once; the motion itself may be iteratively revised during generation and validation. In our implementation, both the accepted reference motion and the initial reward remain fixed throughout policy training.

To provide the one-shot baseline with the same initial learning guidance as
{\method}, we use the same reference-generation procedure, semantic checks, and
initial reward. In addition to kinematic and physical validity, the generated
motion is checked against the requested task semantics, style, and required
action phases before training. The two methods therefore begin from the same
reference and reward; they differ in whether these quantities remain fixed
after observing policy execution.

Concretely, One-shot Motion is the terminal policy obtained from the first
formal trial of {\method}. The trial
uses $4{,}096$ parallel environments and a cumulative budget of at most
$3{,}000$ PPO iterations. Once this trial is complete, the reference motion
and reward are fixed, and no execution-grounded revision is performed.
Ours may instead use subsequent trials to revise the motion, reward, or both
according to policy execution feedback.

This comparison focuses on the effect of keeping the generated learning guidance open to revision after policy execution. Each refinement trial retrains the policy from scratch without reusing checkpoints from previous trials, so improvements across rounds cannot be attributed to warm-starting or accumulated policy optimization. One-shot Motion trains once with the accepted reference, whereas {\method} retrains under the same initialization protocol after each revision. If {\method} terminates after the first trial, the two methods coincide.

\subsection{Eureka-style Reward Refinement}
\label{app:eureka-implementation}

\paragraph{Eureka-style. }Our Eureka-style baseline instantiates the reward-search and reward-reflection
procedure of Eureka~\citep{ma2024eureka} within our common RL interface. For
each robot and task, the agent first constructs the task environment and an
evaluator. Before reward search begins, the observation interface, action
mapping, reset and termination rules, PPO configuration, and evaluation
interface are fixed. Reward design is then the component revised throughout
the search. The agent may modify reward terms, functional forms, weights,
and temperature parameters, but does not use reference motions or framewise
tracking objectives. This provides an iterative reward-only counterpart to
{\method}.

We run two reward-search rounds with two candidates per round. Each candidate
is trained from scratch using $4{,}096$ parallel environments for
$2{,}250$ PPO iterations. The two candidates within a round are proposed
before either training result is observed. After training, GPT receives the
resulting reward-component statistics and task-evaluation feedback, compares
the candidates, and determines how to construct the next round of reward
programs. The second round therefore builds on the reward code and execution
feedback from the first round, while policy parameters, optimizer state, and
observation-normalization statistics are reinitialized for every candidate.
Candidate comparison and reward revision are performed by the agent rather
than by a hand-crafted selection rule.

We adapt Eureka's reward-refinement procedure to the same agent, simulator,
and RL stack used by the other methods. Whereas the original Eureka
experiments use GPT-4-0314, pre-existing benchmark environments and fitness
functions, IsaacGym, and substantially broader reward search, our
implementation uses GPT-6 Astra, Isaac Lab, RSL-RL, and a two-round search
with two candidates per round. The search breadth is chosen to match the
formal training allocation used by GPT and {\method} while preserving Eureka's
core loop of reward generation, policy training, execution feedback, and
reward revision.

\subsection{Training Budgets and Final-Policy Evaluation}
\label{app:method-budgets}

All methods use the actor architecture and robot-specific simulation timing
described in Appendix~\ref{app:environment}, with $24$ policy steps collected
per environment per PPO iteration. We control training resources by matching
the maximum formal PPO allocation of GPT, {\method}, and Eureka-style:
\begin{equation}
3 \times 3{,}000
= 2 \times 2 \times 2{,}250
= 9{,}000
\quad \text{PPO iterations per task}.
\end{equation}
With $4{,}096$ parallel environments, each method therefore receives at most
$884{,}736{,}000$ environment transitions per task. One-shot Motion consists
only of the first training trial and is limited to $3{,}000$ PPO iterations,
or $294{,}912{,}000$ transitions. A method may consume less than its maximum
allocation when training terminates before exhausting the budget.

For GPT and {\method}, evaluation uses the terminal policy from the last completed
formal trial, while One-shot Motion uses the terminal policy from the first
trial. For Eureka-style, GPT compares the trained reward candidates using the
feedback available within the reward-search procedure and determines which
candidate to retain. We do not introduce an additional hand-crafted ranking
or selection criterion for choosing among Eureka-style candidates.

All submitted policies are subsequently evaluated using the same external
user-study rubric and reference-independent physical metrics. These final
evaluation measurements are used only for cross-method comparison and do not
feed back into policy or reward selection.

The original Eureka protocol performs additional multi-seed PPO retraining
after reward search. In our comparison, the policy returned by the budgeted
search is used directly for final evaluation, without additional post-search
training. This keeps the formal PPO allocation comparable across GPT,
Eureka-style, and {\method}.

\section{User Study Protocol}
\label{app:protocols}
\label{app:user-study}

Because the compared methods may construct different reference motions and
learning objectives, method-specific quantities such as tracking errors or
training rewards are not directly comparable across methods. We therefore
evaluate the final learned policies under a common behavioral criterion:
whether the executed behavior fulfills the natural-language instruction while
maintaining satisfactory execution quality. The user study operationalizes
this criterion through a shared rating rubric applied identically to all
methods.

\subsection{Participant Instructions and Rating Rubric}

Before rating the policies, participants read a common set of instructions
defining the evaluation criteria and the five-point rating scale. For each
assigned task, participants are shown the natural-language task description
together with four policy videos presented in randomized order. Each video is
rated independently on an integer scale from 1 to 5; because the study
evaluates each policy rather than requiring a ranking among methods, equal
ratings are allowed.

The rubric asks participants to jointly consider two aspects of the final behavior. First, \emph{task-semantic fulfillment} measures whether the policy realizes the behavior specified by the instruction, including the requested action, style, and temporal phases when applicable. Because a task may admit multiple valid motion realizations, participants are instructed not to prefer one realization merely for appearing closer to a particular motion once the task requirements are satisfied. Second, \emph{execution quality} captures properties of the realized motion, including smoothness, stability, coordination, recovery, and recognizable stepping when relevant. These aspects are combined into a single holistic rating: once task semantics are satisfied, differences in rating should primarily reflect execution quality rather than preference for a particular realization. Table~\ref{tab:app-rating-scale} summarizes the five rating anchors.

\begin{table}[tbp]
  \centering
  \caption{Five-point rating rubric used in the user study.}
  \label{tab:app-rating-scale}
  \begin{tabular}{@{}lp{0.87\linewidth}@{}}
    \toprule
    Score & Description \\
    \midrule
    1 & The task is largely unfulfilled, or the motion has severe problems. \\
    2 & Few task requirements are fulfilled, or motion problems are substantial. \\
    3 & The task is partially or mostly fulfilled, with noticeable shortcomings. \\
    4 & The task is fulfilled with generally stable motion and minor problems. \\
    5 & The task is fulfilled with smooth, coordinated, and stable motion. \\
    \bottomrule
  \end{tabular}
\end{table}

\subsection{Video Presentation and Study Interface}
\label{app:study-interface}

The study contains 76 policy videos, corresponding to four methods (Eureka-style, One-shot motion, GPT and {\method}) evaluated
on each of the 19 tasks. All videos depict executions of the learned policies
and are presented at 1280$\times$720 resolution and 30 frames per second.
Within each task, the camera configuration and playback rate are held fixed
across methods, while each video preserves the duration of its corresponding
policy rollout. 

For each task, all four methods are presented with the same natural-language
task description and are identified only by anonymous labels (A--D).

\subsection{Participants, Task Assignment, and Response Filtering}
\label{app:study-responses}
The study includes 57 participants. Participants are assigned subsets of the 19 tasks, so the number of valid raters varies across tasks. Within each task, however, the same participants evaluate all four methods. This paired presentation keeps the rater population fixed within each method comparison and avoids confounding method differences with differences in task-specific rater composition.

Incomplete or otherwise excluded questionnaires are omitted from the analysis. The questionnaire system additionally applies a three-second minimum completion-time filter and rejects submissions below this threshold. All reported user-study statistics are computed only from the remaining valid responses.

\section{Metric Definitions and Aggregation}
\label{app:metrics}

\subsection{Human-Rating Statistics}
\label{app:rating-statistics}

For robot $r$, method $m$, and task $t$, let $\mathcal{R}_{rmt}$ index
the valid individual ratings and let $n_{rmt}=|\mathcal{R}_{rmt}|$.
Validity follows Appendix~\ref{app:study-responses}: only submitted,
non-excluded questionnaires are included. Counts refer to rating entries;
one participant may contribute ratings to several tasks. The task-level
mean and sample standard deviation are
\begin{equation}
\bar{s}_{rmt}
=
\frac{1}{n_{rmt}}
\sum_{i \in \mathcal{R}_{rmt}} s_i,
\qquad
\sigma_{rmt}
=
\sqrt{
\frac{1}{n_{rmt}-1}
\sum_{i \in \mathcal{R}_{rmt}}
\left(s_i-\bar{s}_{rmt}\right)^2
},
\end{equation}
where the sample standard deviation is defined for $n_{rmt}>1$.

For a task group $g$ (walking or non-walking), let $\mathcal{T}_{rg}$
denote its tasks for robot $r$ and let
$n_{rmg}=\sum_{t\in\mathcal{T}_{rg}}n_{rmt}$. We pool all valid ratings
within each robot, method, and group:
\begin{equation}
\bar{s}_{rmg}
=
\frac{1}{n_{rmg}}
\sum_{t\in\mathcal{T}_{rg}}\sum_{i\in\mathcal{R}_{rmt}}s_i
=
\frac{\sum_{t\in\mathcal{T}_{rg}}n_{rmt}\bar{s}_{rmt}}
     {\sum_{t\in\mathcal{T}_{rg}}n_{rmt}}.
\end{equation}
Thus, each task contributes in proportion to its number of valid ratings.
The symbols $n_W$ and $n_S$ in the rating table denote $n_{rmg}$ for
walking and non-walking skill tasks, respectively.

The corresponding sample standard deviation is computed from the same
pooled ratings:
\begin{equation}
\sigma_{rmg}
=\sqrt{\frac{1}{n_{rmg}-1}
\sum_{t\in\mathcal{T}_{rg}}\sum_{i\in\mathcal{R}_{rmt}}
\left(s_i-\bar{s}_{rmg}\right)^2}.
\end{equation}
It describes variation among individual ratings in that cell.

For the robot-level score, $\mathcal{T}_r$ includes all tasks of robot $r$:
\begin{equation}
\bar{s}_{rm}
=
\frac{\sum_{t\in\mathcal{T}_r}n_{rmt}\bar{s}_{rmt}}
     {\sum_{t\in\mathcal{T}_r}n_{rmt}}.
\end{equation}
The Overall column in Appendix~\ref{app:full-results} uses this mean and
the sample standard deviation of all ratings for that robot and method.
Combining the walking and non-walking groups therefore weights their means
by $n_W$ and $n_S$.

The cross-robot Overall and User Rating columns in the main paper give equal
weight to the four unrounded robot-level means:
\begin{equation}
\bar{s}_{m}
=
\frac{1}{4}\sum_r \bar{s}_{rm}.
\end{equation}

This two-stage rule is shared by all user-rating comparisons: individual
ratings have equal weight within each robot, and the four robots have equal
weight in the cross-robot comparison. Means and sample standard deviations
are reported to two decimal places. Boldface denotes the largest unrounded
mean among the displayed methods, including ties.

\subsection{Physical Metrics}
\label{app:physical-metrics}

Physical metrics are computed from the evaluation rollout of each
method--task pair.

Let $J$ denote the number of named joints recorded in the evaluation trace,
including stationary joints, and let
$a_t,q_t\in\mathbb{R}^{J}$ denote the commanded joint-position targets and
measured joint positions, respectively, at sample $t$. Both are expressed in
radians. Commands are evaluated after the task-specific action mapping, such
that the metric operates on the joint targets actually sent to the controller. $N$ denotes the number of remaining valid
samples. All finite differences use consecutive samples without padding or
filtering. The recorded joint counts are $J=12$ for Go2, $14$ for MicroDuck,
$15$ for Ostrich, and $8$ for Gorilla.

\paragraph{Yaw Err.}
For a walking trajectory with $N$ samples, let $\psi_t$ be the body yaw obtained
from the recorded world-frame quaternion in $(w,x,y,z)$ order. The requested
straight-walking direction is world $+X$, with $\psi^{\star}=0$. We compute
\begin{equation}
  E_{\mathrm{yaw}}=\frac{1}{N}\sum_{t=1}^{N}
  \left|\operatorname{atan2}\!\left(
  \sin(\psi_t-\psi^{\star}),\cos(\psi_t-\psi^{\star})\right)\right|.
\end{equation}
The result is measured in radians and quantifies the mean absolute
deviation from the desired heading.

\paragraph{Second-order Action Variation (Action Var.).}
Writing $\delta^2 a_t=a_t-2a_{t-1}+a_{t-2}$ for the second difference
of the commanded joint-position targets, we compute
\begin{equation}
  E_{\mathrm{action}}=
  \frac{1}{(N-2)J}\sum_{t=3}^{N}\|\delta^2 a_t\|_2^2.
\end{equation}
This metric measures changes in successive command increments, with
lower values indicating smoother command evolution. It is averaged
over time and all recorded joints, including stationary joints.
The result is in rad$^2$.

\paragraph{Joint acceleration RMS (Joint Acc.).}
With measurement interval $\Delta t$, we estimate
\begin{equation}
  \widehat{\ddot q}_t=
  \frac{q_t-2q_{t-1}+q_{t-2}}{\Delta t^2},
  \qquad
  E_{\mathrm{acc}}=
  \sqrt{\frac{1}{(N-2)J}\sum_{t=3}^{N}
  \|\widehat{\ddot q}_t\|_2^2}.
\end{equation}
The result is in rad/s$^2$ and characterizes the magnitude of changes
in joint velocity. Squared accelerations are averaged over time and
joints before taking the square root. Acceleration is estimated from
positions at the metric sampling interval.

\paragraph{Contact-conditioned slip RMS (Foot-slip).}
Let $v^{xy}_{t,f}$ be the world-horizontal velocity of support point $f$ and
$c_{t,f}$ its contact indicator. We compute
\begin{equation}
  E_{\mathrm{slip}}=
  \sqrt{\frac{\sum_{t,f}c_{t,f}\|v^{xy}_{t,f}\|_2^2}
  {\sum_{t,f}c_{t,f}}}.
\end{equation}
The result is in m/s and measures horizontal support-point motion
during detected contact. Gorilla includes both hands and feet as
potential support points. Contact is detected when the recorded net
contact-force magnitude on the corresponding body exceeds 10 N for
Go2, 0.5 N for MicroDuck, or 0.3 N for Ostrich and Gorilla.

Go2 uses calf-attached foot-center proxies, MicroDuck uses configured
ankle support points, Ostrich uses foot-link origins, and Gorilla
uses the local hand and foot support points defined by the task.

\subsection{Sampling and Evaluation Windows}

The evaluation protocol specifies a control frequency of 50 Hz,
corresponding to a control interval of 0.02 s.
Metrics are computed over the complete recorded task window at the policy's
native playback speed and task duration.

To preserve a common evaluation window, automatic environment resets are
disabled during metric collection. Native termination events are still
recorded, while the rollout continues until the end of the task window. This
allows the physical metrics to characterize the full execution shown for
evaluation, including behavior following an early failure.

\subsection{Table Aggregation and Reporting}
\label{app:physical-aggregation}

Each physical metric is first computed independently for each task. Task-level
values are then averaged equally over the applicable tasks for each robot,
and the resulting robot-level values are averaged equally across the four
embodiments. Yaw error and contact-conditioned slip are reported over the
seven walking tasks, whereas action variation and joint acceleration are
computed over all 19 tasks.

Tasks are aggregated only after their individual metrics have been computed,
so differences in task duration do not change their relative weight. For RMS
metrics, the square root is taken at the task level before averaging across
tasks and robots.

The resulting physical metrics characterize complementary properties of
policy execution. Lower action second difference and joint acceleration
indicate smoother commanded and realized joint trajectories, respectively,
while yaw error measures heading deviation and contact-conditioned slip
measures support-point motion during contact. These diagnostics are reported
together with the human ratings, which assess whether the requested
behavior is successfully realized.

Writing $\mathcal{T}_{rk}$ for the applicable tasks of robot $r$ for metric
$k$, and $E_{rmtk}$ for a task-level measurement, the reported means are
\begin{equation}
  \bar{E}_{rmk}=\frac{1}{|\mathcal{T}_{rk}|}
  \sum_{t\in\mathcal{T}_{rk}}E_{rmtk},
  \qquad
  \bar{E}_{mk}=\frac{1}{4}\sum_r\bar{E}_{rmk}.
\end{equation}

A dash denotes a metric that is not applicable to a given task; in
particular, yaw and slip are evaluated only on walking tasks. All applicable
rollouts are retained in the aggregation, including unsuccessful executions.
Undefined measurements are reported as such.

\section{Detailed Evaluation Results}
\label{app:full-results}

This appendix provides a more detailed breakdown of the three-method comparison
between Eureka-style, One-shot, and {\method}. We first disaggregate the user ratings
by robot and task group to identify where refinement changes the final behavior,
and then report physical execution metrics by robot. Together, these results
complement the aggregate results in the main paper by showing where
refinement yields substantial gains.

\subsection{User Ratings by Robot and Task Group}
\label{app:detailed-user-ratings}

Table~\ref{tab:app-study-groups} separates the user ratings into walking and
non-walking tasks for each robot. Relative to One-shot, {\method} obtains a higher
overall mean rating on Go2, MicroDuck, and Gorilla, while One-shot remains
slightly higher on Ostrich. The task-group breakdown further localizes these
differences. On Go2, the overall gain is driven by walking behaviors, whose mean
changes from 3.00 to 3.76. MicroDuck improves in both groups, from 2.82 to 4.00
for walking and from 3.60 to 4.05 for non-walking tasks. On Gorilla, walking
remains comparable (3.71 versus 3.79), whereas the non-walking mean increases
from 2.95 to 3.76. On Ostrich, the non-walking means are tied at 4.25, while the
walking mean changes slightly from 3.71 to 3.57.

Importantly, One-shot is the first policy-training trial of our pipeline. This comparison therefore evaluates the benefit of continuing execution-grounded refinement from the same initial motion and reward. The clearest gains occur when the first-trial policy
is relatively weak: MicroDuck improves in both task groups, Gorilla improves
substantially on non-walking tasks, and Go2 walking also shows a marked gain.
When the first trial already performs strongly, refinement changes the ratings
only modestly. This pattern suggests that refinement is most useful when the initially
generated guidance does not yet translate into satisfactory policy execution.
Rather than requiring every additional trial to improve every task, the loop
keeps the motion and reward revisable when execution reveals a mismatch between
the generated guidance and what the policy can successfully realize.

\begin{table}[!htbp]
  \centering
  \caption{User ratings by robot and task group ($\uparrow$), reported as
  mean $\pm$ sample standard deviation. $n_W$ and $n_S$ denote the numbers of
  valid individual ratings per method for walking and non-walking tasks,
  respectively, and are not counts of distinct participants. Bold indicates
  the highest unrounded mean among the displayed methods within each robot and
  column, including ties.}
  \label{tab:app-study-groups}
  \begin{tabular*}{\linewidth}{@{\extracolsep{\fill}}llrcrcc@{}}
    \toprule
    Robot & Method & $n_W$ & Walking & $n_S$ & Non-walking & Overall \\
    \midrule
    Go2 & Eureka-style & 29 & $3.00\pm1.54$ & 43 & $2.65\pm1.38$ & $2.79\pm1.44$ \\
     & One-shot & 29 & $3.00\pm1.44$ & 43 & $\mathbf{4.56}\pm0.63$ & $3.93\pm1.28$ \\
     & {\method} (Ours) & 29 & $\mathbf{3.76}\pm0.91$ & 43 & $4.33\pm0.78$ & $\mathbf{4.10}\pm0.87$ \\
    \midrule
    MicroDuck & Eureka-style & 28 & $2.64\pm1.34$ & 43 & $1.60\pm0.85$ & $2.01\pm1.18$ \\
     & One-shot & 28 & $2.82\pm1.09$ & 43 & $3.60\pm1.29$ & $3.30\pm1.27$ \\
     & {\method} (Ours) & 28 & $\mathbf{4.00}\pm1.02$ & 43 & $\mathbf{4.05}\pm0.92$ & $\mathbf{4.03}\pm0.96$ \\
    \midrule
    Ostrich & Eureka-style & 28 & $2.07\pm1.09$ & 44 & $1.95\pm1.12$ & $2.00\pm1.10$ \\
     & One-shot & 28 & $\mathbf{3.71}\pm0.85$ & 44 & $\mathbf{4.25}\pm0.75$ & $\mathbf{4.04}\pm0.83$ \\
     & {\method} (Ours) & 28 & $3.57\pm0.96$ & 44 & $\mathbf{4.25}\pm0.84$ & $3.99\pm0.94$ \\
    \midrule
    Gorilla & Eureka-style & 14 & $1.14\pm0.36$ & 42 & $1.74\pm0.80$ & $1.59\pm0.76$ \\
     & One-shot & 14 & $3.71\pm0.61$ & 42 & $2.95\pm1.41$ & $3.14\pm1.30$ \\
     & {\method} (Ours) & 14 & $\mathbf{3.79}\pm0.58$ & 42 & $\mathbf{3.76}\pm0.93$ & $\mathbf{3.77}\pm0.85$ \\
    \bottomrule
  \end{tabular*}
\end{table}

\subsection{Physical Execution Metrics by Robot}
\label{app:detailed-physical-results}

Table~\ref{tab:app-robot-metrics} reports the physical execution metrics under the same three-method comparison. {\method} achieves the lowest Foot-slip on all four embodiments and the lowest Action Var. and Joint Acc. on Go2, MicroDuck, and Ostrich, while One-shot performs best on the latter two metrics for Gorilla. Directional accuracy is more embodiment dependent: Eureka-style achieves the lowest Yaw MAE on Go2, One-shot on Ostrich, and {\method} on MicroDuck and Gorilla. Overall, {\method} achieves the best value in 12 of the 16 robot--metric comparisons, with particularly consistent gains in support quality and execution smoothness.

This pattern complements the user-study results. The physical metrics characterize specific aspects of execution, including directional accuracy, command smoothness, realized joint motion, and contact slip, whereas the user ratings assess whether the policy realizes the requested behavior as a whole. A favorable value on an individual physical metric therefore does not necessarily imply successful task execution. All method--task outcomes are retained in the analysis, with each metric aggregated only over the tasks for which it is defined.

\begin{table}[!htbp]
  \centering

  \caption{Physical execution metrics by robot ($\downarrow$). Yaw error and
  slip are averaged over walking tasks, whereas action variation and joint
  acceleration are averaged over all tasks. Values are rounded to three
  significant digits. Bold indicates the minimum among the displayed methods
  within each robot and metric, including ties. Action var. denotes
  second-order action variation normalized by the number of named joints.}
  \label{tab:app-robot-metrics}
  \begin{tabular*}{\linewidth}{@{\extracolsep{\fill}}llrrrr@{}}
    \toprule
    Robot & Method & Yaw Err. & Action Var. & Joint Acc. & Foot-slip \\
          &        & (rad) & (rad$^2$) & (rad/s$^2$) & (m/s) \\
    \midrule
    Go2 & Eureka-style & $\mathbf{0.00807}$ & $0.0531$ & $72.7$ & $0.148$ \\
     & One-shot & $0.063$ & $0.00307$ & $23.3$ & $0.2$ \\
     & {\method} (Ours) & $0.0135$ & $\mathbf{0.000174}$ & $\mathbf{13.9}$ & $\mathbf{0.0354}$ \\
    \midrule
    MicroDuck & Eureka-style & $0.0779$ & $0.034$ & $122$ & $0.0564$ \\
     & One-shot & $0.139$ & $0.0222$ & $77.9$ & $0.0674$ \\
     & {\method} (Ours) & $\mathbf{0.00435}$ & $\mathbf{0.0102}$ & $\mathbf{46.4}$ & $\mathbf{0.0104}$ \\
    \midrule
    Ostrich & Eureka-style & $0.032$ & $0.0585$ & $144$ & $0.0582$ \\
     & One-shot & $\mathbf{0.00656}$ & $0.00106$ & $20.7$ & $0.00592$ \\
     & {\method} (Ours) & $0.0121$ & $\mathbf{0.000712}$ & $\mathbf{16.9}$ & $\mathbf{0.00324}$ \\
    \midrule
    Gorilla & Eureka-style & $0.02$ & $0.186$ & $119$ & $0.223$ \\
     & One-shot & $0.0255$ & $\mathbf{0.00023}$ & $\mathbf{4.88}$ & $0.0133$ \\
     & {\method} (Ours) & $\mathbf{0.0035}$ & $0.000482$ & $10.8$ & $\mathbf{0.00353}$ \\
    \bottomrule
  \end{tabular*}
\end{table}

\subsection{Types of Motion and Reward Revisions}
\label{app:refinement}

The main-paper Q2 analysis records 15 motion revisions across 11 of the
19 evaluated tasks. These counts characterize how often motion revision is
used. The observed
changes fall into four recurring types: \emph{posture and limb trajectory},
including body height, leg folding, and swing clearance; \emph{timing and
phase structure}, including swing duration and limb ordering; \emph{support
transfer}, including preservation of stance contacts and body shifts before
limb lift; and \emph{whole-body coordination}, which couples the moving limb
with the torso and the remaining support limbs. A single revision may involve
multiple types, so these categories are not treated as mutually exclusive
frequency counts.

These revision types clarify what keeping motion in the loop contributes to
the refinement process. Motion revision directly changes the behavioral
realization used to guide policy learning---for example, its posture, timing,
support pattern, or whole-body coordination---while reward revision changes
the optimization objective used to learn that behavior. Both remain constrained
by the original task semantics, and they can be applied jointly when execution
feedback indicates that changes to both the behavioral guidance and the learning
objective are needed.

\section{Unguided GPT Analysis}
\label{app:unguided-gpt}

\subsection{Experimental Setup}
\label{app:gpt-setup}

The agent is provided with the robot, task specification, and
the same actor template available to our framework, but receives
no explicit instruction to generate a reference motion or to follow the
motion--reward refinement procedure. The agent is therefore left to determine
how to formulate the learning problem, including whether to introduce a motion
reference and whether to revise it in response to policy execution. This
ablation evaluates the contribution of explicitly prescribing
execution-grounded motion--reward refinement, while holding access to the
underlying learning tools and actor template fixed. For evaluation, we use the
policy from the agent's final formal trial, following the selection rule in
Appendix~\ref{app:implementation}.

As reported in the main paper, unguided GPT introduces reference motions in 16 of 19 tasks and subsequently revises them in seven. Thus, motion guidance can emerge even without being explicitly prescribed. This ablation therefore tests a stronger question than whether GPT can discover motion guidance at all: whether it can use execution feedback to refine that guidance effectively. We evaluate the resulting policies using the same user-rating protocol and physical execution metrics as for {\method}, separating the emergence of a refinement strategy from its effectiveness in downstream policy learning.

\subsection{User Ratings on Walk and Non-walk Tasks}
\label{app:gpt-user-ratings}

Table~\ref{tab:app-gpt-study-groups} disaggregates the user-study ratings for GPT and {\method} into walking and non-walking tasks for each robot. {\method} obtains a higher mean rating in both task groups across all four embodiments. This consistent pattern shows that the higher user ratings are not driven by a single task category, but are observed across both walking and non-walking behaviors. The scores follow the aggregation procedure defined in Appendix~\ref{app:metrics}.

\begin{table}[t]
  \centering
  \caption{User ratings on walking and non-walking tasks ($\uparrow$), reported as mean $\pm$ sample standard deviation. Bold indicates the higher mean between GPT and {\method} within each robot and task group.}
  \label{tab:app-gpt-study-groups}
  \begin{tabular*}{\linewidth}{@{\extracolsep{\fill}}llcc@{}}
    \toprule
    Robot & Method & Walking & Non-walking \\
    \midrule
    Go2 & GPT & $3.14\pm1.36$ & $3.86\pm1.15$ \\
     & {\method} (Ours) & $\mathbf{3.76}\pm0.91$ & $\mathbf{4.33}\pm0.78$ \\
    \midrule
    MicroDuck & GPT & $3.61\pm1.13$ & $3.91\pm1.06$ \\
     & {\method} (Ours) & $\mathbf{4.00}\pm1.02$ & $\mathbf{4.05}\pm0.92$ \\
    \midrule
    Ostrich & GPT & $3.21\pm1.10$ & $4.05\pm0.83$ \\
     & {\method} (Ours) & $\mathbf{3.57}\pm0.96$ & $\mathbf{4.25}\pm0.84$ \\
    \midrule
    Gorilla & GPT & $2.57\pm0.85$ & $3.64\pm0.93$ \\
     & {\method} (Ours) & $\mathbf{3.79}\pm0.58$ & $\mathbf{3.76}\pm0.93$ \\
    \bottomrule
  \end{tabular*}
\end{table}

\subsection{Physical Metrics by Robot}
\label{app:gpt-physical-results}

Table~\ref{tab:app-gpt-robot-metrics} reports the physical execution metrics for GPT and {\method}, aggregated by robot according to the procedure in Appendix~\ref{app:metrics}. Lower values indicate better execution for all reported metrics.

Across the four embodiments and four reported metrics, {\method} achieves the lower value in 10 of 16 robot--metric comparisons, including lower slip on all four embodiments. Although several individual metrics favor GPT, the overall pattern favors {\method} across the measured execution properties. Together with the user ratings in Table~\ref{tab:app-gpt-study-groups}, these results provide complementary evidence for the benefit of explicitly prescribed execution-grounded refinement across both task-level behavioral assessment and physical execution.

\begin{table}[t]
  \centering

  \caption{Physical execution metrics of GPT and {\method} across robot embodiments ($\downarrow$). Yaw error and slip are averaged over walking tasks, whereas action variation and joint acceleration are averaged over all tasks. Bold indicates the lower value. }
  \label{tab:app-gpt-robot-metrics}
  \begin{tabular*}{\linewidth}{@{\extracolsep{\fill}}llrrrr@{}}
    \toprule
    Robot & Method & Yaw Err. & Action Var. & Joint Acc. & Foot-slip  \\
          &        & (rad) & (rad$^2$) & (rad/s$^2$) & (m/s) \\
    \midrule
    Go2 & GPT & $\mathbf{0.00746}$ & $0.0698$ & $68.9$ & $0.0905$ \\
     & {\method} (Ours) & $0.0135$ & $\mathbf{0.000174}$ & $\mathbf{13.9}$ & $\mathbf{0.0354}$ \\
    \midrule
    MicroDuck & GPT & $0.0133$ & $\mathbf{0.00106}$ & $\mathbf{22.3}$ & $0.0241$ \\
     & {\method} (Ours) & $\mathbf{0.00435}$ & $0.0102$ & $46.4$ & $\mathbf{0.0104}$ \\
    \midrule
    Ostrich & GPT & $0.0143$ & $\mathbf{0.00054}$ & $17.7$ & $0.0355$ \\
     & {\method} (Ours) & $\mathbf{0.0121}$ & $0.000712$ & $\mathbf{16.9}$ & $\mathbf{0.00324}$ \\
    \midrule
    Gorilla & GPT & $0.0337$ & $\mathbf{0.0000328}$ & $\mathbf{7.83}$ & $0.0175$ \\
     & {\method} (Ours) & $\mathbf{0.0035}$ & $0.000482$ & $10.8$ & $\mathbf{0.00353}$ \\
    \bottomrule
  \end{tabular*}
\end{table}

\begin{figure}[!t]
  \centering
  \IfFileExists{figures/walk_lie.pdf}{%
    \includegraphics[width=\linewidth]{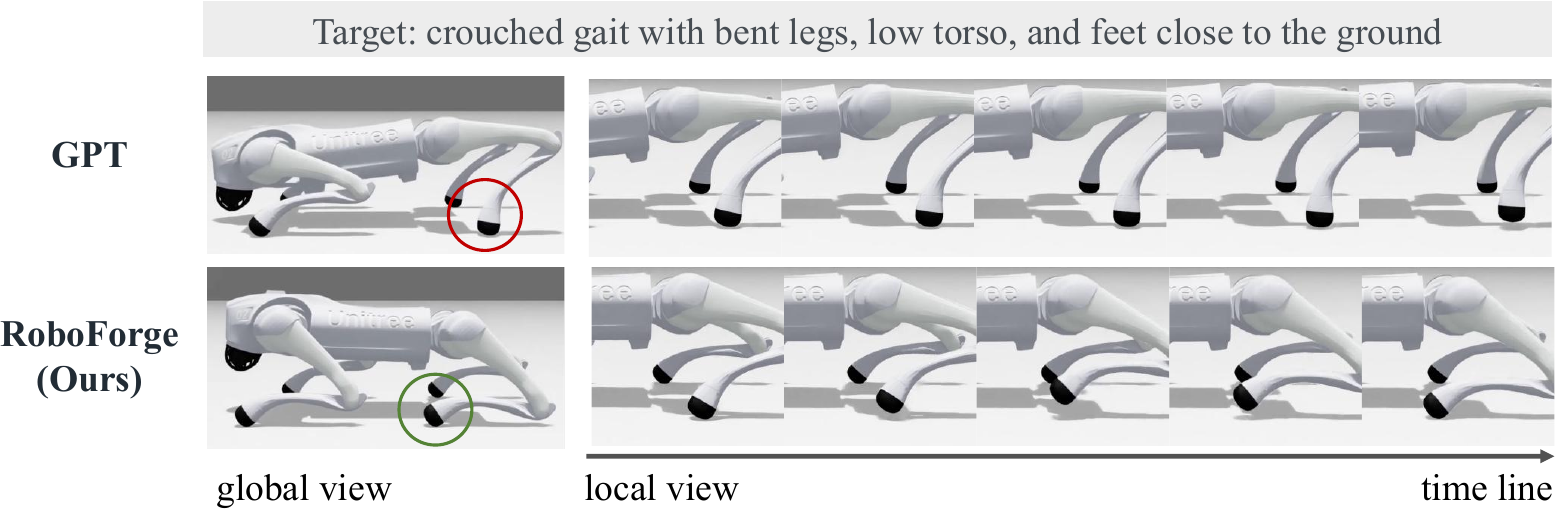}%
  }{%
    \figureplaceholder{0.14\textheight}{Insert the final-policy comparison for Go2 walk\_lie: GPT and {\method}, using matched views and temporal order to show stepping, crouched posture, and support coordination.}%
  }
  \caption{Final learned policy executions for the Go2 low-walk task. Figure~\ref{fig:q3} shows the corresponding reference-motion revisions, whereas this figure shows the final policies learned from those refined motions. {\method} more clearly realizes the requested crouched gait, with distinct stepping, sustained low posture, and coordinated support transfer across the gait cycle.}
  \label{fig:app-walk-lie}
\end{figure}

\subsection{From Motion Refinement to Policy Execution: A Low-Walk Case}
\label{app:gpt-low-walk}

The low-walk task provides a concrete case study connecting motion refinement to downstream policy learning. Figure~\ref{fig:q3} shows the corresponding reference-motion refinements. Both GPT and {\method} identify foot sliding as a failure mode and increase swing-foot clearance, while our refinement further adjusts body motion and the supporting limbs to coordinate the swing-leg revision with posture maintenance and support transfer. The reference motions themselves, however, are intermediate learning guidance rather than the final output of the system. The central question is therefore whether these revisions provide more effective guidance for learning the requested behavior.

Figure~\ref{fig:app-walk-lie} shows the corresponding final policy executions. Although both methods revise the reference in response to the same initial sliding failure, the learned policies realize the requested behavior differently. The policy trained with our refined motion more clearly combines distinct stepping with maintenance of the crouched posture and coordinated support transfer, whereas increasing swing-foot clearance alone does not fully resolve the whole-body gait. The two figures therefore connect the motion revision to its downstream consequence: our refinement does not merely produce a different reference trajectory, but produces guidance that leads to a more complete realization of the requested behavior by the learned policy. 

This case illustrates the role of keeping motion in the execution-grounded refinement loop. Rollout feedback determines how the intermediate motion guidance should be revised, while the resulting policy execution determines whether that revision was useful for learning. Rather than treating the quality of the reference motion as an end in itself, our procedure treats motion as revisable learning guidance whose utility is evaluated through downstream policy behavior.

\section{Robot Generation Checks}
\label{app:rg2-protocol}

\subsection{Evaluation Setup}
\label{app:rg2-setup}
\label{app:rg2-status}

We evaluate four gorilla-inspired robots generated by GPT6 Astra,
GPT5.6 Sol, GPT~5.5, and GPT5.6 Terra.
The two tasks are forward swinging (\texttt{flat\_palm\_swing})
and hand waving (\texttt{flat\_palm\_wave}).
GPT6 Astra executes the check programs for all four designs;
the programs compute geometric, motion, and engineering measurements
and determine the corresponding Pass/Fail outcomes.

The checks are grouped into \textbf{design and assembly},
\textbf{motion and task execution}, and \textbf{engineering}.
The three groups contain 10, 7, and 6 independent checks,
respectively. Each check is counted once and all counts are unweighted.
\emph{Static} checks use geometry or model properties,
\emph{reference} checks use the prescribed motion,
and \emph{execution} checks use recorded robot motion.

\subsection{Check Definitions}
\label{app:rg2-catalog}
\label{app:rg2-motion}

\paragraph{Design and assembly.}
These checks progress from individual parts to the complete assembly.
\emph{Solid validity} verifies that each STEP part forms a valid
single solid with positive volume.
\emph{Manufacturing mesh} checks that each printable mesh is closed,
has positive volume, and is positioned correctly on the print bed.
\emph{Export volume fidelity} verifies that geometry is preserved
from the source model to the STEP and STL exports.
\emph{BOM completeness} checks that printed parts, motors, files,
and quantities are consistent.
\emph{Interface geometry} examines mounting holes, tool openings,
and supporting material around the motor interface.
\emph{Print envelope} checks whether each part fits the available
build volume in its printing orientation.
\emph{Fit tolerance} evaluates worst-case clearance at mating
interfaces under dimensional variation.
\emph{Assembly and tool access} checks whether components can be
inserted and tools can reach the required fasteners without
interference.
\emph{Fastener interfaces} checks screw length, grip, engagement,
bottoming, and head clearance at the mechanical connections.
\emph{Static assembly clearance} checks the neutral full assembly
for unintended penetration between components.

\paragraph{Motion and task execution.}
These checks evaluate the articulated model and the robot behavior
throughout the two tasks.
\emph{Mechanical model} verifies joint axes, order, limits,
masses, inertias, the free base, and the eight actuated joints.
\emph{Forward swing} checks palm support, rear-foot lift,
forward body motion, and stable landing.
\emph{Hand wave} checks left-hand lift, support from the other
three pads, shoulder and elbow motion, and stable landing.
\emph{Motion interference} checks the CAD geometry for collisions
through the recorded reference and executed poses.
\emph{Simulation tracking} measures joint and root-position errors
between the executed and reference motions.
\emph{Actuator limits} checks peak torque, saturation,
and joint-limit reaction torque.
\emph{Continuous clearance} checks whether positive geometric
clearance remains throughout the motion after applying the specified
manufacturing, assembly, calibration, tracking, elastic, thermal,
model, and safety allowances.

\paragraph{Engineering checks.}
These checks evaluate load capacity and mechanical margins.
\emph{Local structural margin} evaluates critical sections under
the loads recorded during motion.
\emph{Mass consistency} reconciles material estimates, link masses,
and CAD and simulation totals.
\emph{Fastener length chain} verifies that the selected screws
span the clamped parts, engage the threads, and avoid bottoming out.
\emph{Joint-limit margin} measures the distance between executed
joint angles and the URDF travel limits.
\emph{Load-path strength} evaluates load-bearing parts and
connections under support, landing, and emergency-stopping loads.
\emph{Cable routing} checks clearance, bending radius, and twist
throughout the joint range.

\subsection{Check Results}

\begin{longtable}{@{}llcccc@{}}
\caption{Pass/Fail results for all robot-generation checks.
The category counts in the main paper can be recomputed directly
from this table.}
\label{tab:rg2-all-check-results}
\label{tab:rg2-all-results}
\label{tab:rg2-motion-results}
\label{tab:rg2-engineering-results}\\
\toprule
ID & Check &
\shortstack{GPT6\\Astra} &
\shortstack{GPT5.6\\Sol} &
\shortstack{GPT5.6\\Terra} &
GPT 5.5 \\
\midrule
\endfirsthead

\toprule
ID & Check &
\shortstack{GPT6\\Astra} &
\shortstack{GPT5.6\\Sol} &
\shortstack{GPT5.6\\Terra} &
GPT 5.5 \\
\midrule
\endhead

\bottomrule
\endfoot

\multicolumn{6}{l}{\textit{Design and assembly}} \\
DA01 & Solid validity            & Pass & Pass & Pass & Fail \\
DA02 & Manufacturing mesh        & Pass & Fail & Fail & Fail \\
DA03 & Export volume fidelity    & Pass & Pass & Pass & Pass \\
DA04 & BOM completeness          & Pass & Pass & Pass & Pass \\
DA05 & Interface geometry        & Pass & Fail & Fail & Fail \\
DA06 & Print envelope            & Pass & Pass & Pass & Pass \\
A05  & Fit tolerance             & Fail & Fail & Fail & Fail \\
A06  & Assembly and tool access  & Fail & Fail & Fail & Fail \\
A07  & Fastener interfaces       & Fail & Fail & Fail & Fail \\
A09  & Static assembly clearance & Fail & Fail & Fail & Fail \\

\midrule
\multicolumn{6}{l}{\textit{Motion and task execution}} \\
DB01 & Mechanical model          & Pass & Pass & Pass & Pass \\
DB02 & Forward swing             & Pass & Fail & Fail & Fail \\
DB03 & Hand wave                 & Pass & Fail & Fail & Fail \\
DB04 & Motion interference       & Pass & Fail & Fail & Fail \\
DB05 & Simulation tracking       & Fail & Fail & Fail & Fail \\
DB06 & Actuator limits           & Pass & Fail & Fail & Fail \\
B05  & Continuous clearance      & Fail & Fail & Fail & Fail \\

\midrule
\multicolumn{6}{l}{\textit{Engineering checks}} \\
DC01 & Local structural margin   & Pass & Fail & Fail & Fail \\
DC02 & Mass consistency          & Pass & Fail & Fail & Fail \\
DC03 & Fastener length chain     & Pass & Fail & Fail & Fail \\
DC04 & Joint-limit margin        & Pass & Fail & Pass & Fail \\
C02  & Load-path strength        & Fail & Fail & Fail & Fail \\
C05  & Cable routing             & Fail & Fail & Fail & Fail \\
\end{longtable}

\subsection{Evaluation Parameters}
\label{app:rg2-assumptions}

\paragraph{Fit and assembly.}
Printed dimensions use a $\pm0.2$\,mm size tolerance,
hole position uses a 0.2\,mm radial bound,
purchased-part dimensions use $\pm0.05$\,mm,
and assembly position uses a 0.1\,mm bound.
For the M2 interfaces, the worst-case radial clearance is
\[
\frac{d_h-0.2-d_s-0.05}{2}-0.2-0.1,
\]
which must remain positive.
Tool access is evaluated with a radius-2\,mm,
length-30\,mm bit and a radius-6\,mm,
length-60\,mm handle approaching coaxially with the screw.

\paragraph{Task completion.}
For forward swinging, both front palms must support the robot
at normal force $>0.01$\,N each while both rear feet remain
more than 2\,mm above the ground for at least 0.5\,s.
The body must advance by at least 10\,mm during the same interval.
For hand waving, the left palm must remain more than 50\,mm
above the ground while the other three pads each provide
normal support $>0.01$\,N for at least 1\,s.
The left shoulder and elbow excursions must each reach
at least 0.1\,rad.
Both tasks require a final stable landing lasting at least 1\,s,
with four-pad support, upright body orientation, low linear
and angular velocity, and no non-pad ground contact.

\paragraph{Tracking and actuation.}
Simulation tracking requires a maximum joint error
$\leq5^\circ$ and a maximum root-position error $\leq5$\,mm
over both motions.
Actuator checks require peak torque $\leq0.100001$\,N\,m,
saturation fraction $\leq0.001$,
and joint-limit reaction torque $\leq0.001$\,N\,m.

\paragraph{Continuous clearance.}
The translational clearance budget includes
0.4\,mm for manufacturing,
0.1\,mm for assembly,
0.2\,mm for elasticity,
0.05\,mm for thermal effects,
0.05\,mm for model error,
and 0.2\,mm for safety.
Angular allowances are $0.5^\circ$ for calibration
and $5^\circ$ for tracking.
For a point at farthest radius $r$ from a joint axis,
an angular allowance $\theta$ contributes
$2r\sin(\theta/2)$ to the displacement budget.

\paragraph{Structural loads.}
The local structural check uses an allowable stress of 8\,MPa,
a stress-concentration factor of 2,
and an additional reserve factor of 3.
The local margin is
\[
\frac{8}{6\sigma_{\rm eq}},
\]
and must be at least one.
Load-path checks cover support, landing, and emergency stopping.
Landing uses velocity $0.5$\,m/s and stopping distance
$0.005$\,m, giving
\[
F_{\rm avg}=m\left(g+\frac{v^2}{2s}\right).
\]
Emergency stopping uses the recorded velocity over 0.1\,s.
The allowable stresses are 8\,MPa for bulk material,
4\,MPa transverse to print layers,
and 4\,MPa in connection shear.

\paragraph{Mass accounting.}
Mass accounting reconciles material estimates,
link assignments, and whole-robot CAD and simulation masses.
Material densities are 1.24, 1.27, and 1.21\,g/cm$^3$
for PLA, PETG, and TPU, respectively.
Each motor is modeled as 18\,g,
with 16.8\,g assigned to the stator and 1.2\,g to the rotor.
The controller, battery, horn supports,
and fasteners/wiring contribute 45\,g, 54\,g, 16\,g,
and 25\,g, respectively.

\paragraph{Fastener length chains.}
For the horn connection, screw engagement must satisfy
\[
L_h-g_h\in(0,3]\ {\rm mm}.
\]
For the front and rear motor-case connections,
the engagement lengths are
\[
L_c-g_c-3.5
\quad\text{and}\quad
L_c-g_c-4.5,
\]
respectively, and both must lie strictly between
0 and 15\,mm.

\paragraph{Joint travel and cable routing.}
The minimum distance to the configured joint limits
must remain at least $1^\circ$ over all recorded frames
of both tasks.
Cable routing uses a 180\,mm cable with a 1.5\,mm
envelope radius.
The route must maintain a bend radius of at least 18\,mm
and a twist no greater than $90^\circ$ per 100\,mm.

\end{document}